\documentclass{article}

\usepackage{microtype}
\usepackage{graphicx}
\usepackage{subfigure}
\usepackage{amsmath}
\usepackage{booktabs} 
\usepackage{amssymb}
\usepackage{multirow}
\usepackage{tabularx}
\usepackage{enumitem}
\usepackage{physics}
\usepackage{amsthm}
\usepackage{algorithm}
\DeclareMathOperator*{\argmax}{arg\,max}
\DeclareMathOperator*{\argmin}{arg\,min}
\usepackage{lscape}

\usepackage{tikz}
\usetikzlibrary{arrows,shapes,positioning,shadows,trees}

\tikzset{
  basic/.style  = {draw, text width=2cm, drop shadow, font=\sffamily, rectangle},
  root/.style   = {basic, rounded corners=2pt, thin, align=center,
                   fill=green!30},
  level 2/.style = {basic, rounded corners=6pt, thin,align=center, fill=green!60,
                   text width=11em},
  level 3/.style = {basic, thin, align=left, fill=pink!60, text width=12.5em}
}

\usepackage{hyperref}

\usepackage[accepted]{icml2019}

\icmltitlerunning{Adversarial Attacks and Defenses in Images, Graphs and Text: A Review}

\begin{document}

\twocolumn[
\icmltitle{Adversarial Attacks and Defenses in Images, Graphs and Text: A Review}



\icmlsetsymbol{equal}{*}

\begin{icmlauthorlist}
\icmlauthor{Han Xu}{msu1}
\icmlauthor{Yao Ma}{msu2}
\icmlauthor{Haochen Liu}{msu3}
\icmlauthor{Debayan Deb}{msu4}
\icmlauthor{Hui Liu}{msu5}
\icmlauthor{Jiliang Tang}{msu6}
\icmlauthor{Anil K. Jain}{msu7}

\end{icmlauthorlist}

\icmlaffiliation{msu1}{Michigan State University, \{xuhan1@msu.edu\} $~~~~~~~~~~~~~~~~~$}
\icmlaffiliation{msu2}{Michigan State University, \{mayao4@msu.edu\}$~~~~~~~~~~~~~~~~~$}
\icmlaffiliation{msu3}{Michigan State University, \{liuhaoc1@msu.edu\}$~~~~~~~~~~~~~~~~~$}
\icmlaffiliation{msu4}{Michigan State University, \{debdebay@msu.edu\}$~~~~~~~~~~~~~~~~~$}
\icmlaffiliation{msu5}{Michigan State University, \{liuhui7@msu.edu\}$~~~~~~~~~~~~~~~~~$}
\icmlaffiliation{msu6}{Michigan State University, \{tangjili@msu.edu\}$~~~~~~~~~~~~~~~~~$}
\icmlaffiliation{msu7}{Michigan State University, \{jain@egr.msu.edu\}$~~~~~~~~~~~~~~~~~$}

\icmlkeywords{Machine Learning, ICML}

\vskip 0.3in
]



\printAffiliationsAndNotice{} 

\begin{abstract}
Deep neural networks (DNN) have achieved unprecedented success in numerous machine learning tasks in various domains. However, the existence of adversarial examples has raised concerns about applying deep learning to safety-critical applications. As a result, we have witnessed increasing interests in studying attack and defense mechanisms for DNN models on different data types, such as images, graphs and text. Thus, it is necessary to provide a systematic and comprehensive overview of the main threats of attacks and the success  of corresponding countermeasures. In this survey, we review the state of the art algorithms for generating adversarial examples and the countermeasures against adversarial examples, for the three popular data types, i.e., images, graphs and text.
\end{abstract}

\section{Introduction}

Deep neural networks have become increasingly popular and successful in many machine learning tasks. They have been deployed in different recognition problems in the domains of images, graphs, text and speech, with remarkable success. In the image recognition domain, they are able to recognize objects with near-human level accuracy \citep{krizhevsky2012imagenet,he2016deep}. They are also used in speech recognition \citep{hinton2012deep}, natural language processing \citep{hochreiter1997long} and for playing games \citep{44806}. 

Because of these accomplishments, deep learning techniques are also applied in safety-critical tasks. For example, in autonomous vehicles, deep convolutional neural networks (CNNs) are used to recognize road signs \citep{cirecsan2012multi}. The machine learning technique used here is required to be highly accurate, stable and reliable. But, what if the CNN model fails to recognize the ``STOP'' sign by the roadside and the vehicle keeps going? It will be a dangerous situation. Similarly, in financial fraud detection systems, companies frequently use graph convolutional networks (GCNs) \citep{kipf2016semi} to decide whether their customers are trustworthy or not. If there are fraudsters disguising their personal identity information to evade the company's detection, it will cause a huge loss to the company.  Therefore, the safety issues of deep neural networks have become a major concern.

In recent years, many works \citep{szegedy2013intriguing,goodfellow2014explaining,he2016deep} have shown that DNN models are vulnerable to adversarial examples, which can be formally defined as --  \textit{``Adversarial examples are inputs to machine learning models that an attacker intentionally designed to cause the model to make mistakes.''} In the image classification domain, these adversarial examples are intentionally synthesized images which look almost exactly the same as the original images (see figure \ref{fig:panda}), but can mislead the classifier to provide wrong prediction outputs. For a well-trained DNN image classifier on the MNIST dataset, almost all the digit samples can be attacked by an imperceptible perturbation, added on the original image. Meanwhile, in other application domains involving graphs, text or audio, similar adversarial attacking schemes also exist to confuse deep learning models. For example, perturbing only a couple of edges can mislead graph neural networks \citep{zugner2018adversarial}, and inserting typos to a sentence can fool text classification or dialogue systems \citep{ebrahimi2017hotflip}. As a result, the existence of adversarial examples in all application fields has cautioned researchers against directly adopting DNNs in safety-critical machine learning tasks. 

To deal with the threat of adversarial examples, studies have been published with the aim of finding countermeasures to protect deep neural networks. These approaches can be roughly categorized to three main types: (a) \textit{Gradient Masking} \citep{papernot2016distillation,athalye2018obfuscated}: since most attacking algorithms are based on the gradient information of the classifiers, masking or obfuscating the gradients will confuse the attack mechanisms. (b) \textit{Robust Optimization} \citep{madry2017towards,kurakin2016adversarial1}: these studies show how to train a robust classifier that can correctly classify the adversarial examples. (c) \textit{Adversary Detection} \citep{carlini2017adversarial,xu2017feature}: the approaches attempt to check whether a sample is benign or adversarial before feeding it to the deep learning models. It can be seen as a method of guarding against adversarial examples. These methods improve DNN's resistance to adversarial examples.

In addition to building safe and reliable DNN models, studying adversarial examples and their countermeasures is also beneficial for us to understand the nature of DNNs and consequently improve them. For example, adversarial perturbations are perceptually indistinguishable to human eyes but can evade DNN's detection. This suggests that the DNN's predictive approach does not align with human reasoning. There are works \citep{goodfellow2014explaining,ilyas2019adversarial} to explain and interpret the existence of adversarial examples of DNNs, which can help us gain more insight into DNN models.

In this review, we aim to summarize and discuss the main studies dealing with adversarial examples and their countermeasures. We provide a systematic and comprehensive review on the start-of-the-art algorithms from images, graphs and text domain, which gives an overview of the main techniques and contributions to adversarial attacks and defenses. The main structure of this survey is as follows: 

In Section \ref{def}, we introduce some important definitions and concepts which are frequently used in adversarial attacks and their defenses. It also gives a basic taxonomy of the types of attacks and defenses. In Section \ref{attack} and Section \ref{defend}, we discuss main attack and defense techniques in the image classification scenario. We use Section \ref{explain} to briefly introduce some studies which try to explain the phenomenon of adversarial examples. Section \ref{graph} and Section \ref{nlp} review the studies in graph and text data, respectively. 
\section{Definitions and Notations}\label{def}

In this section, we give a brief introduction to the key components of model attacks and defenses. We hope that our explanations can help our audience to understand the main components of the related works on adversarial attacks and their countermeasures. By answering the following questions, we define the main terminology: 
\begin{itemize}[topsep=0pt,itemsep = 0pt]
    \item \textit{Adversary's Goal}\hspace{0.2cm} (\ref{goal})\\
    What is the goal or purpose of the attacker? Does he want to misguide the classifier's decision on one sample, or influence the overall performance of the classifier?
    \item \textit{Adversary's Knowledge}\hspace{0.2cm}(\ref{cap})\\
    What information is available to the attacker? Does he know the classifier's structure, its parameters or the training set used for classifier training? 
    \item \textit{Victim Models}\hspace{0.2cm}(\ref{vic})\\
    What kind of deep learning models do adversaries usually attack? Why are adversaries interested in attacking these models?
    \item \textit{Security Evaluation}\hspace{0.2cm}(\ref{rob})\\
    How can we evaluate the safety of a victim model when faced with adversarial examples? What is the relationship and difference between these security metrics and other model goodness metrics, such as accuracy or risks?
\end{itemize}

\subsection{Threat Model}

\subsubsection{Adversary's Goal} \label{goal}
\begin{itemize}
    \item Poisoning Attack vs Evasion Attack\\
    Poisoning attacks refer to the attacking algorithms that allow an attacker to insert/modify several fake samples into the training database of a DNN algorithm. These fake samples can cause failures of the trained classifier. They can result in the poor accuracy \citep{biggio2012poisoning}, or wrong prediction on some given test samples \citep{zugner2018adversarial}. This type of attacks frequently appears in the situation where the adversary has access to the training database. For example, web-based repositories and ``honeypots'' often collect malware examples for training, which provides an opportunity for adversaries to poison the data. 
    
    In evasion attacks, the classifiers are fixed and usually have good performance on benign testing samples. The adversaries do not have authority to change the classifier or its parameters, but they craft some fake samples that the classifier cannot recognize. In other words, the adversaries generate some fraudulent examples to evade detection by the classifier. For example, in autonomous driving vehicles, sticking a few pieces of tapes on the stop signs can confuse the vehicle's road sign recognizer \citep{eykholt2017robust}.  
    
    
    \item Targeted Attack vs Non-Targeted Attack\\
    In targeted attack, when the victim sample $(x,y)$ is given, where $x$ is feature vector and $y\in \mathcal{Y}$ is the ground truth label of $x$, the adversary aims to induce the classifier to give a specific label $t\in \mathcal{Y}$ to the perturbed sample $x'$. For example, a fraudster is likely to attack a financial company's credit evaluation model to disguise himself as a highly credible client of this company.
    
    If there is no specified target label $t$ for the victim sample $x$, the attack is called non-targeted attack. The adversary only wants the classifier to predict incorrectly.
\end{itemize}

\subsubsection{Adversary's Knowledge}\label{cap}
\begin{itemize}
    \item White-Box Attack \\
    In a white-box setting, the adversary has access to all the information of the target neural network, including its architecture, parameters, gradients, etc. The adversary can make full use of the network information to carefully craft adversarial examples. White-box attacks have been extensively studied because the disclosure of model architecture and parameters helps people understand the weakness of DNN models clearly and it can be analyzed mathematically. As stated by \citep{tramer2017ensemble}, security against white-box attacks is the property that we desire ML models to have.
    
    \item Black-Box Attack \\
     In a black-box attack setting, the inner configuration of DNN models is unavailable to adversaries. Adversaries can only feed the input data and query the outputs of the models. They usually attack the models by keeping feeding samples to the box and observing the output to exploit the model's input-output relationship, and identity its weakness. Compared to white-box attacks, black-box attacks are more practical in applications because model designers usually do not open source their model parameters for proprietary reasons. 
    \item Semi-white (Gray) Box Attack \\
    In a semi-white box or gray box attack setting, the attacker trains a generative model for producing adversarial examples in a white-box setting. Once the generative model is trained, the attacker does not need victim model anymore, and can craft adversarial examples in a black-box setting. 

\end{itemize}

\subsubsection{Victim Models} \label{vic}
We briefly summarize the machine learning models which are susceptible to adversarial examples, and some popular deep learning architectures used in image, graph and text data domains. In our review, we mainly discuss studies of adversarial examples for deep neural networks.

\begin{enumerate}[label=\Alph*]
    \item \textit{Conventional Machine Learning Models}
    
    For conventional machine learning tools, there is a long history of studying safety issues. \citet{biggio2013evasion} attack SVM classifiers and fully-connected shallow neural networks for the MNIST dataset. \citet{barreno2010security} examine the security of SpamBayes, a Bayesian method based spam detection software. In \citep{dalvi2004adversarial}, the security of Naive Bayes classifiers is checked. Many of these ideas and strategies have been adopted in the study of adversarial attacks in deep neural networks. 

    \item \textit{Deep Neural Networks}\\
    Different from traditional machine learning techniques which require domain knowledge and manual feature engineering, DNNs are end-to-end learning algorithms. The models use raw data directly as input to the model, and learn objects' underlying structures and attributes. The end-to-end architecture of DNNs makes it easy for adversaries to exploit their weakness, and generate high-quality deceptive inputs (adversarial examples).  Moreover, because of the implicit nature of DNNs, some of their properties are still not well understood or interpretable. Therefore, studying the security issues of DNN models is necessary. Next, we'll briefly introduce some popular victim deep learning models which are used as ``benchmark'' models in attack/defense studies.
   
    \begin{enumerate}
        \item \textit{Fully-Connected Neural Networks}\\
        Fully-connected neural networks (FC) are composed of layers of artificial neurons. In each layer, the neurons take the input from previous layers, process it with the activation function and send it to the next layer; the input of first layer is sample $x$, and the (softmax) output of last layer is the score $F(x)$. An $m$-layer fully connected neural network can be formed as:
        \begin{equation*}\hspace{0.2cm}
            z^{(0)} = x;~~~~
            z^{(l+1)} = \sigma(W^l z^{l}+b^l).
        \end{equation*}
        One thing to note is that, the back-propagation algorithm helps calculate $\pdv{F(x;\theta)}{\theta}$, which makes gradient descent effective in learning parameters. In adversarial learning, back-propagation also facilitates the calculation of the term: $\pdv{F(x;\theta)}{x}$, representing the output's response to a change in input. This term is widely used in the studies to craft adversarial examples.
        \item \textit{Convolutional Neural Networks}\\
        In computer vision tasks, Convolutional Neural Networks \citep{krizhevsky2012imagenet} is one of the most widely used models. CNN models aggregate the local features from the image to learn the representations of image objects. CNN models can be viewed as a sparse-version of fully connected neural networks: most of the weights between layers are zero. Its training algorithm or gradients calculation can also be inherited from fully connected neural networks.
        
        \item \textit{Graph Convolutional Networks}\\
        The work of \citep{kipf2016semi} introduces the graph convolutional networks, which later became a popular node classification model for graph data. The idea of graph convolutional networks is similar to CNN: it aggregates the information from neighbor nodes to learn representations for each node $v$, and outputs the score $F(v,X)$ for prediction:
        \begin{equation*}\hspace{0.2cm}
            H^{(0)} = X; ~~~
            H^{(l+1)} = \sigma(\hat{A} H^{(l)}W^l).
        \end{equation*}
        where $X$ denotes the input graph's feature matrix, and $\hat{A}$ depends on graph degree matrix and adjacency matrix.
        
        \item \textit{Recurrent Neural Networks}\\
        Recurrent Neural Networks are very useful for tackling sequential data. As a result, they are widely used in natural language processing. The RNN models, especially LSTM\citep{hochreiter1997long}, are able to store the previous time information in memory, and exploit useful information from previous sequence for next-step prediction.
    \end{enumerate}
        
\end{enumerate}

\subsection{Security Evaluation}\label{rob}
We also need to evaluate the model's resistance to adversarial examples. ``Robustness'' and ``Adversarial Risk'' are two terms used to describe this resistance of DNN models to one single sample, and the total population, respectively. 

\subsubsection{Robustness}\label{mip}

\newtheorem{definition}{Definition}
\begin{definition}
  \textbf{(minimal perturbation)}:
  Given the classifier $F$ and data $(x,y)$, the adversarial perturbation has the least norm (the most unnoticeable perturbation): 
   \begin{equation*}
        \delta_{min} = \argmin_{\delta}||\delta|| ~~~\text{subject to}~~~ F(x+\delta) \neq y.
  \end{equation*}
  Here, $||\cdot||$ usually refers to $l_p$ norm.
\end{definition}
\begin{definition}
  \textbf{(robustness)}:
  The norm of minimal perturbation:
  \begin{equation*}
      r(x,F) = ||\delta_{min}||.
  \end{equation*}
\end{definition}
\begin{definition} \textbf{(global robustness)}:
The expectation of robustness over the whole population $D$:
    \begin{equation*}
        \rho(F) = \displaystyle \mathop{\mathbb{E}}_{x\sim \mathcal{D}}~ r(x,F).
  \end{equation*}
\end{definition}
The minimal perturbation can find the adversarial example which is most similar to $x$ under the model $F$. Therefore, the larger $r(x,F)$ or $\rho(F)$ is, the adversary needs to sacrifice more similarity to generate adversarial samples, implying that the classifier $F$ is more robust or safe.

\subsubsection{Adversarial Risk (Loss)}\label{adr}


\begin{definition}
\textbf{(Most-Adversarial Example)}
  Given the classifier $F$ and data $x$, the sample $x_{adv}$ with the largest loss value in $x$'s $\epsilon$-neighbor ball: 
  \begin{equation*}
        x_{adv} = \argmax_{x'}\mathcal{L}(x',F)~~~\text{subject to}~~~ ||x'-x||\leq \epsilon.
  \end{equation*}
\end{definition}
\begin{definition}
\textbf{(Adversarial Loss)}: The loss value for the most-adversarial example:
\begin{equation*}
    \mathcal{L}_{adv} (x) = \mathcal{L}(x_{adv}) = \max_{||x'-x||<\epsilon}~\mathcal{L}  (\theta,x',y)
\end{equation*}
\end{definition}

\begin{definition} 
\textbf{(global adversarial loss)}:
The expectation of the loss value on $x_{adv}$ over the data distribution $\mathcal{D}$:
\begin{equation} \label{eq:ar}
    \mathcal{R}_{adv}(F) = \displaystyle \mathop{\mathbb{E}}_{x\sim \mathcal{D}}~ \max_{||x'-x||<\epsilon}~\mathcal{L}  (\theta,x',y)
  \end{equation}
\end{definition}

The most-adversarial example is the point where the model is most likely to be fooled in the neighborhood of $x$. A lower loss value $\mathcal{L}_{adv}$ indicates a more robust model $F$.

\subsubsection{Adversarial Risk vs Risk}

The definition of \textit{Adversarial Risk} is drawn from the definition of classifier risk (empirical risk): 
\begin{equation*}
     \mathcal{R}(F) = \displaystyle \mathop{\mathbb{E}}_{x\sim \mathcal{D}}~ \mathcal{L}  (\theta,x,y)
\end{equation*}
Risk studies a classifier's performance on samples from natural distribution $\mathcal{D}$. Whereas, adversarial risk from Equation (\ref{eq:ar}) studies a classifier's performance on adversarial example $x'$. It is important to note that $x'$ may not necessarily follow the distribution $\mathcal{D}$. Thus, the studies on adversarial examples are different from these on model generalization. Moreover, a number of studies reported the relation between these two properties \citep{tsipras2018robustness,su2018robustness,stutz2019disentangling, zhang2019theoretically}. From our clarification, we hope that our audience get the difference and relation between risk and adversarial risk, and the importance of studying adversarial countermeasures.

\subsection{Notations}
With the aforementioned definitions, Table \ref{tab:notation} lists the notations which will be used in the subsequent sections.
\begin{table}[ht]
    \centering
   \begin{tabular}{ | c | p{6cm} |}
    \hline
    Notations& Description\\
    \hline
    $x$ & Victim data sample\\
    $x'$ & Perturbed data sample\\
    $\delta$ & Perturbation\\
    $B_\epsilon(x)$ & $l_p$-distance neighbor ball around $x$ with radius $\epsilon$\\
    $\mathcal{D}$ & Natural data distribution\\
    $||\cdot||_p$ & $l_p$ norm\\
    \hline
    $y$ & Sample $x$'s ground truth label\\
    $t$ & Target label $t$\\
    $\mathcal{Y}$ & Set of possible labels. Usually we assume there are $m$ labels\\
    \hline
    $C$ & Classifier whose output is a label: $C(x)=y$\\
    $F$ & DNN model which outputs a score vector: $F(x) \in [0,1]^m$\\
    $Z$ & Logits: last layer outputs before softmax: $F(x) = softmax(Z(x))$\\
    $\sigma$ & Activation function used in neural networks\\
    \hline
    
    $\theta$  & Parameters of the model $F$\\
    $\mathcal{L}$ & Loss function for training. We simplify $\mathcal{L}(F(x),y)$ in the form $\mathcal{L}(\theta,x,y)$.\\
    \hline
    \end{tabular}
    \caption{Notations.}
    \label{tab:notation}
\end{table}

\section{Generating Adversarial Examples} \label{attack}
In this section, we introduce main methods for generating adversarial examples in the image classification domain. Studying adversarial examples in the image domain is considered to be essential because: (a) perceptual similarity between fake and benign images is intuitive to observers, and (b) image data and image classifiers have simpler structure than other domains, like graph or audio. Thus, many studies concentrate on attacking image classifiers as a standard case. In this section,  we assume that the image classifiers refer to fully connected neural networks and Convolutional Neural Networks \citep{krizhevsky2012imagenet}. The most common datasets used in these studies include (1) handwritten letter images dataset MNIST, (2) CIFAR10 object dataset and (3) ImageNet \citep{deng2009imagenet}.  Next, we go through some main methods used to generate adversarial image examples for evasion attack (white-box, black-box, grey-box, physical-world attack), and poisoning attack settings.  Note that we also summarize all the attack methods in Table \ref{tab:attack} in the Appendix A.

\subsection{White-box Attacks}
Generally, in a white-box attack setting, when the classifier $C$ (model $F$) and the victim sample $(x,y)$ are given to the attacker, his goal is to synthesize a fake image $x'$ perceptually similar to original image $x$ but it can mislead the classifier $C$ to give wrong prediction results. It can be formulated as: 
\begin{equation*}
\begin{split}
    &\text{find } x' \text{ satisfying } ||x'-x||\leq\epsilon\\
    &\text{such that } C(x')=t\neq y
\end{split}
\end{equation*}
where $||\cdot||$ measures the dissimilarity between $x'$ and $x$, which is usually $l_p$ norm. Next, we will go through main methods to realize this formulation.

\subsubsection{Biggio's Attack}
In \citep{biggio2013evasion}, adversarial examples are generated on the MNIST dataset targeting conventional machine learning classifiers like SVMs and 3-layer fully-connected neural networks.

\begin{figure}
    \centering
    \includegraphics[width = 75mm]{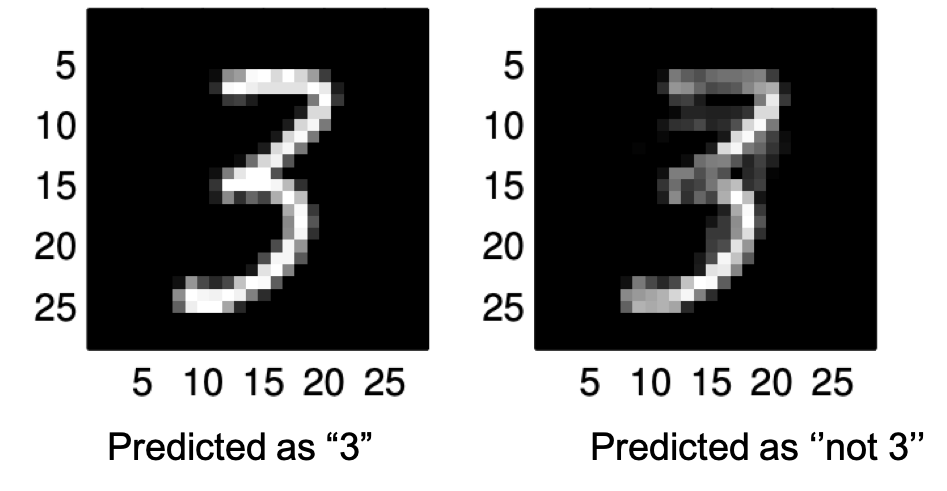}
    \caption{Biggio's attack on SVM classifier for letter recognition. (Image Credit: \citep{biggio2013evasion})}
    \label{fig:4}
\end{figure}

It optimizes the discriminant function to mislead the classifier. For example, on the MNIST dataset, for a linear SVM classifier, its discriminant function $g(x) = \langle w,x\rangle +b$, will mark a sample $x$ with positive value $g(x)>0$ to be in class ``3'', and $x$ with $g(x)\leq 0$ to be in class ``not 3''. An example of this attack is shown in figure \ref{fig:4}.

Suppose we have a sample $x$ which is correctly classified to be ``3''. For this model, Biggio's attack crafts a new example $x'$ to minimize the discriminant value $g(x')$ while keeping $||x'-x||_1$ small. If $g(x')$ is negative, the sample is classified as ``not 3'', but $x'$ is still close to $x$, so the classifier is fooled. The studies about adversarial examples for conventional machine learning models \citep{dalvi2004adversarial,biggio2012poisoning,biggio2013evasion} have inspired investigations on safety issues of deep learning models.

\subsubsection{Szegedy's L-BFGS Attack} \label{sze}
The work of \citep{szegedy2013intriguing} is the first to attack deep neural network image classifiers. They formulate their optimization problem as a search for minimal distorted adversarial example $x'$, with the objective:
\begin{equation}\label{opti1}
    \begin{split}
        &\text{minimize  } ||x-x'||_2^2\\
        &\text{subject to  } C(x')=t \text{  and  } x' \in [0,1]^m
    \end{split}
\end{equation}
The problem is approximately solved by introducing the loss function, which results in the following objective:
\begin{equation*}
    \begin{split}
        &\text{minimize  } c||x-x'||_2^2+\mathcal{L}(\theta,x',t)\\
        &\text{subject to  }  x' \in [0,1]^m
    \end{split}
\end{equation*}
In the optimization objective of this problem, the first term imposes the similarity between $x'$ and $x$. The second term encourages the algorithm to find $x'$ which has a small loss value to label $t$, so the classifier $C$ is very likely to predict $x'$ as $t$. By continuously changing the value of constant $c$, they can find an $x'$ which has minimum distance to $x$, and at the same time fool the classifier $C$. To solve this problem, they implement the L-BFGS \citep{liu1989limited} algorithm.


\subsubsection{Fast Gradient Sign Method} \label{fgsm}
\begin{figure}[t]
    \centering
    \includegraphics[width = 80mm]{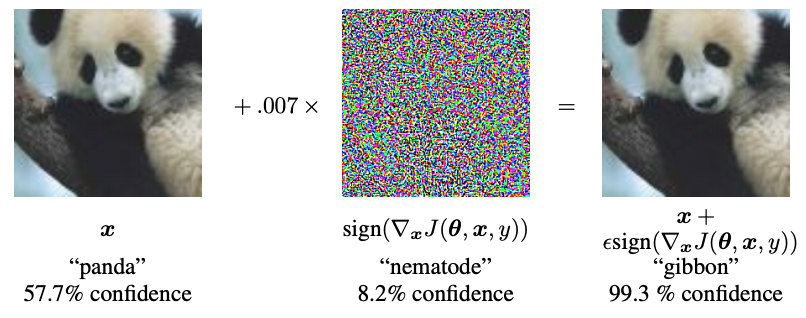}
    \caption{By adding an unnoticeable perturbation, ``panda'' is classified as ``gibbon''. (Image Credit: \citep{goodfellow2014explaining})}
    \label{fig:panda}
\end{figure}

In \citep{goodfellow2014explaining}, an one-step method is introduced to fast generate adversarial examples. The formulation is: 
\begin{equation*}
\begin{split}
    & x' = x + \epsilon \text{sign}(\nabla_x \mathcal{L}(\theta,x,y)) ~~~\text{     non-target}\\
    & x' = x - \epsilon \text{sign}(\nabla_x \mathcal{L}(\theta,x,t))~~~~ \text{   target on t}
\end{split}
\end{equation*}
For a targeted attack setting, this formulation can be seen as a one-step of gradient descent to solve the problem:
\begin{equation}\label{opti2}
    \begin{split}
        &\text{minimize  } \mathcal{L}(\theta,x',t)\\
        &\text{subject to  } ||x'-x||_{\infty}\leq\epsilon \text{  and  } x' \in [0,1]^m
    \end{split}
\end{equation}
The objective function in (\ref{opti2}) searches the point which has the minimum loss value to label $t$ in $x$'s $\epsilon$-neighbor ball, which is the location where model $F$ is most likely to predict it to the target class $t$. In this way, the one-step generated sample $x'$ is also likely to fool the model. An example of FGSM-generated samples on ImageNet is shown in Figure \ref{fig:panda}.

Compared to the iterative attack in Section \ref{sze}, FGSM is fast in generating adversarial examples, because it only involves calculating one back-propagation step. Thus, FGSM addresses the demands of tasks that need to generate a large amount of adversarial examples. For example, adversarial training \citep{kurakin2016adversarial}, uses FGSM to produce adversarial samples for all instances in the training set.

\subsubsection{Deep Fool} \label{deepfool}
\begin{figure} [ht] 
  \centering
  \includegraphics[width=0.8\linewidth]{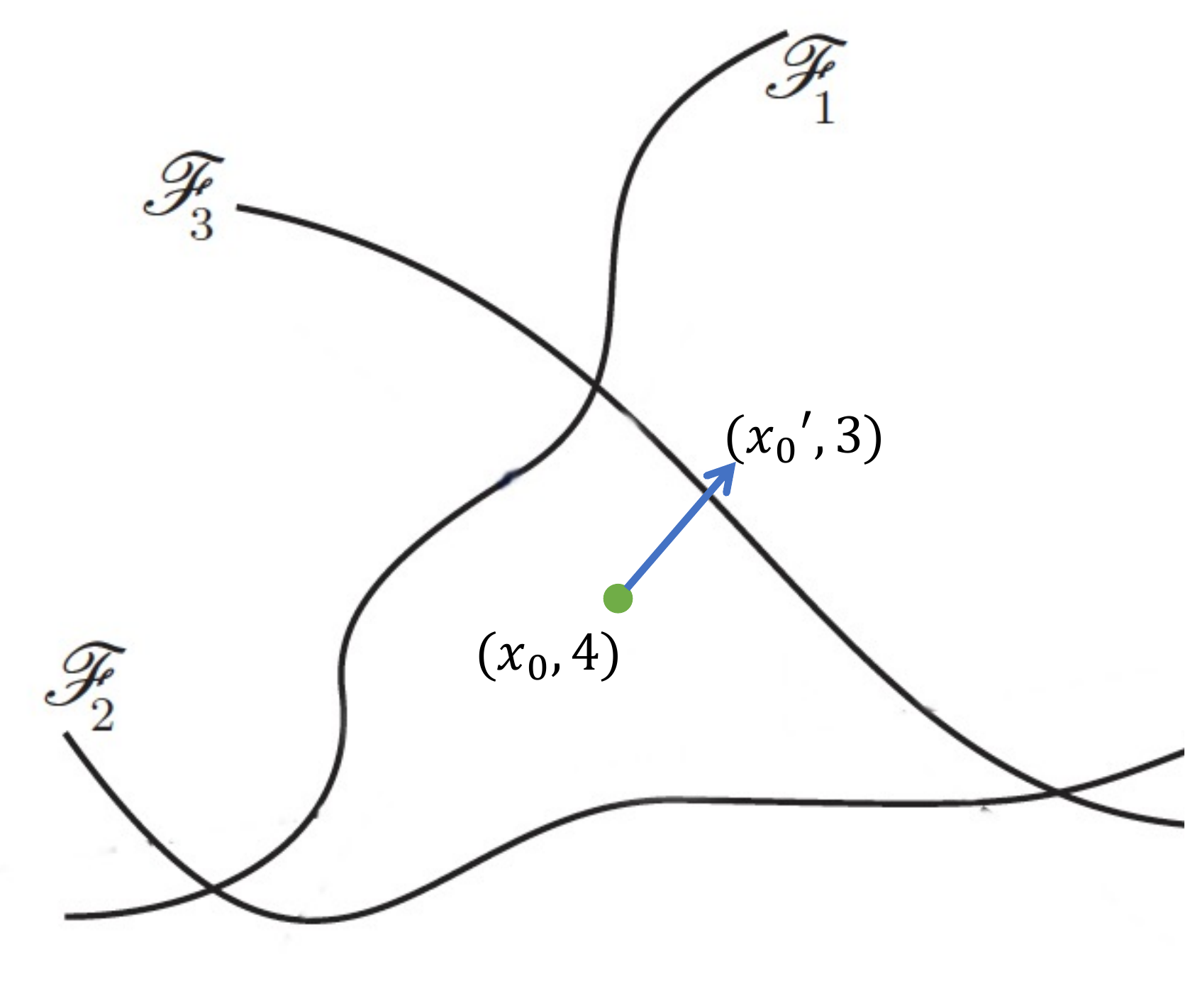}
  \caption{Decision Boundaries: the hyperplane $\mathcal{F}_1$ ($\mathcal{F}_2$ or $\mathcal{F}_3$) separates the data points belonging to class 4 and class 1 (class 2 or 3). The sample $x_0$ crosses the decision boundary $\mathcal{F}_3$, so the perturbed data $x_0'$ is classified as class 3. (Image Credit: \citep{moosavi2016deepfool})}
  \label{fig:deepfool}
\end{figure}

In \cite{moosavi2016deepfool}, the authors study a classifier $F$'s decision boundary around data point $x$. They try to find a path such that $x$ can go beyond the decision boundary, as shown in figure \ref{fig:deepfool}, so that the classifier will give a different prediction for $x$. For example, to attack $x_0$ (true label is digit 4) to digit class 3, the decision boundary is described as $\mathcal{F}_3 =\{z:F(x)_4-F(x)_3=0\}$. We denote $f(x) = F(x)_4-F(x)_3$ for short. In each attacking step, it linearizes the decision boundary hyperplane using Taylor expansion $\mathcal{F}'_3=\{x:f(x)\approx f(x_0)+\langle\nabla_x f(x_0)\cdot (x-x_0)\rangle=0\}$, and calculates the orthogonal vector $\omega$ from $x_0$ to plane $\mathcal{F}'_3$. This vector $\omega$ can be the perturbation that makes $x_0$ go beyond the decision boundary $\mathcal{F}_3$. By moving along the vector $\omega$, the algorithm is able to find the adversarial example $x'_0$ that is classified to class 3.

The experiments of DeepFool \citep{moosavi2016deepfool} show that for common DNN image classifiers, almost all test samples are very close to their decision boundary. For a well-trained LeNet classifier on MNIST dataset, over $90\%$ of test samples can be attacked by small perturbations whose $l_\infty$ norm is below 0.1 where the total range is $[0,1]$. This suggests that the DNN classifiers are not robust to small perturbations.

\subsubsection{Jacobian-based Saliency Map Attack} \label{jsma}
Jacobian-based Saliency Map Attack (JSMA) \citep{papernot2016limitations} introduced a method based on calculating the Jacobian matrix of the score function $F$. It can be viewed as a greedy attack algorithm by iteratively manipulating the pixel which is the most influential to the model output. 

The authors used the Jacobian matrix $J_F(x) = \frac{\partial F(x)}{\partial x} = \{\frac{\partial F_j(x)}{\partial x_i}\}_{i\times j}$ to model $F(x)$'s change in response to the change of its input $x$. For a targeted attack setting where the adversary aims to craft an $x'$ that is classified to the target class $t$, they repeatedly search and manipulate pixel $x_i$ whose increase (decrease) will cause $F_t(x)$ to increase or decrease $\sum_{j \neq t} F_j(x)$. As a result, for $x$, the model will give it the largest score to label $t$.

\subsubsection{Basic Iterative Method (BIM) / Projected Gradient Descent (PGD) Attack}\label{bim}
The Basic Iterative Method was first introduced by \citep{kurakin2016adversarial} and \citep{kurakin2016adversarial1}. It is an iterative version of the one-step attack \textit{FGSM} in Section \ref{fgsm}. In a non-targeted setting, it gives an iterative formulation to craft $x'$:
\begin{equation*}
    \begin{split}
        & x_0 = x\\
        & x^{t+1} = Clip_{x,\epsilon} (x^t + \alpha \text{sign}(\nabla_x \mathcal{L}(\theta, x^t,y)))
    \end{split}
\end{equation*}
Here, $Clip$ denotes the function to project its argument to the surface of $x$'s $\epsilon$-neighbor ball $B_{\epsilon}(x):\{ x' :||x'-x||_\infty\leq \epsilon \}$. The step size $\alpha$ is usually set to be relatively small (e.g. 1 unit of pixel change for each pixel), and step numbers guarantee that the perturbation can reach the border (e.g. $step = \epsilon/\alpha +10$). This iterative attacking method is also known as Projected Gradient Method (PGD) if the algorithm is added by a random initialization on $x$, used in work \citep{madry2017towards}.

This BIM (or PGD) attack heuristically searches the samples $x'$ which have the largest loss value in the $l_\infty$ ball around the original sample $x$. These adversarial examples are called ``most-adversarial'' examples: they are the sample points which are most aggressive and most-likely to fool the classifiers, when the perturbation intensity (its $l_p$ norm) is limited. Finding these adversarial examples is helpful to find the weaknesses of deep learning models.

\subsubsection{Carlini \& Wagner's Attack}\label{cw}
Carlini and Wagner's attack \citep{carlini2017towards} counterattacks the defense strategy \citep{papernot2016distillation} which were shown to be successful against FGSM and L-BFGS attacks. C\&W's attack aims to solve the same problem as defined in L-BFGS attack (section \ref{sze}), namely trying to find the minimally-distorted perturbation (Equation \ref{opti1}). 

The authors address the problem (\ref{opti1}) by instead solving:
\begin{equation*}
    \begin{split}
        &\text{minimize  } ||x-x'||_2^2+c\cdot f(x',t)\\
        &\text{subject to  } x' \in [0,1]^m
    \end{split}
\end{equation*}
where $f$ is defined as $f(x',t) = (\max_{i\neq t}{Z(x')_i}-Z(x')_t)^+$. Minimizing $f(x',t)$ encourages the algorithm to find an $x'$ that has larger score for class $t$ than any other label, so that the classifier will predict $x'$ as class $t$. Next, by applying a line search on constant $c$, we can find the $x'$ that has the least distance to $x$. 

The function $f(x,y)$ can also be viewed as a loss function for data $(x,y)$: it penalizes the situation where there are some labels $i$ with scores $Z(x)_i$ larger than $Z(x)_y$. It can also be called margin loss function.

The only difference between this formulation and the one in L-BFGS attack (section \ref{sze}) is that C\&W's attack uses margin loss $f(x,t)$ instead of cross entropy loss $\mathcal{L}(x,t)$. The benefit of using margin loss is that when $C(x')=t$, the margin loss value $f(x',t)=0$, the algorithm will directly minimize the distance from $x'$ to $x$. This procedure is more efficient for finding the minimally distorted adversarial example. 


The authors claim their attack is one of the strongest attacks, breaking many defense strategies which were shown to be successful. Thus, their attacking method can be used as a benchmark to examine the safety of DNN classifiers or the quality of other adversarial examples.

\subsubsection{Ground Truth Attack}\label{ground}
Attacks and defenses keep improving to defeat each other. In order to end this stalemate, the work of \citep{carlini2017provably} tries to find the ``provable strongest attack''. It can be seen as a method to find the theoretical minimally-distorted adversarial examples.

This attack is based on Reluplex \citep{katz2017reluplex}, an algorithm for verifying the properties of neural networks. It encodes the model parameters $F$ and data $(x,y)$ as the subjects of a linear-like programming system, and then solves the system to check whether there exists an eligible sample $x'$ in $x$'s neighbor $B_{\epsilon}(x)$ that can fool the model. If we keep reducing the radius $\epsilon$ of search region $B_{\epsilon}(x)$ until the system determines that there does not exist such an $x'$ that can fool the model, the last found adversarial example is called the ground truth adversarial example, because it has been proved to have least dissimilarity with $x$.

The ground-truth attack is the first work to seriously calculate the exact robustness (minimal perturbation) of classifiers. However, this method involves using a SMT solver (a complex algorithm to check the satisfiability of a series of theories), which will make it slow and not scalable to large networks. More recent works \citep{tjeng2017evaluating,xiao2018training} have improved the efficiency of the ground-truth attack.

\subsubsection{Other $l_p$ Attacks}
Previous studies are mostly focused on $l_2$ or $l_\infty$ norm-constrained perturbations. However, there are other papers which consider other types of $l_p$ attacks.

(a) \textit{One-pixel Attack}. In~\citep{su2019one}, it studies a similar problem as in Section \ref{sze}, but constrains the perturbation's $l_0$ norm. Constraining $l_0$ norm of the perturbation $x'-x$ will limit the number of pixels that are allowed to be changed. It shows that: on CIFAR10 dataset, for a well-trained CNN classifier (e.g. VGG16, which has 85.5\% accuracy on test data), most of the testing samples (63.5\%) can be attacked by changing the value of only one pixel in a non-targeted setting. This also demonstrates the poor robustness of deep learning models.

(b) \textit{EAD: Elastic-Net Attack}. In~\citep{chen2018ead}, it also studies a similar problem as in Section \ref{sze}, but constrains the perturbations $l_1$ and $l_2$ norm together. As shown in their experimental work \citep{sharma2017attacking}, some strong defense models that aim to reject $l_\infty$ and $l_2$ norm attacks \citep{madry2017towards} are still vulnerable to the $l_1$-based Elastic-Net attack.

\subsubsection{Universal Attack}
Previous methods only consider one specific targeted victim sample $x$. However, the work \citep{moosavi2017universal} devises an algorithm that successfully misleads a classifier's decision on almost all testing images. They try to find a perturbation $\delta$ satisfying:
\begin{enumerate}[topsep=0pt,itemsep=-1ex]
  \item $||\delta||_p \leq \epsilon$
  \item $\underset{x \sim D(x)}{\mathbb{P}}   (C(x+\delta) \neq C(x))\leq 1-\sigma$
\end{enumerate}
This formulation aims to find a perturbation $\delta$ such that the classifier gives wrong decisions on most of the samples. In their experiments, for example, they successfully find a perturbation that can attack 85.4\% of the test samples in the ILSVRC 2012 \citep{ILSVRC15} dataset under a ResNet-152 \citep{he2016deep} classifier. 

The existence of ``universal'' adversarial examples reveals a DNN classifier's inherent weakness on all of the input samples. As claimed in \citep{moosavi2017universal}, it may suggest the property of geometric correlation among the high-dimensional decision boundary of classifiers.

\subsubsection{Spatially Transformed Attack}
\begin{figure} [ht] 
  \centering
  \includegraphics[width=0.9\linewidth]{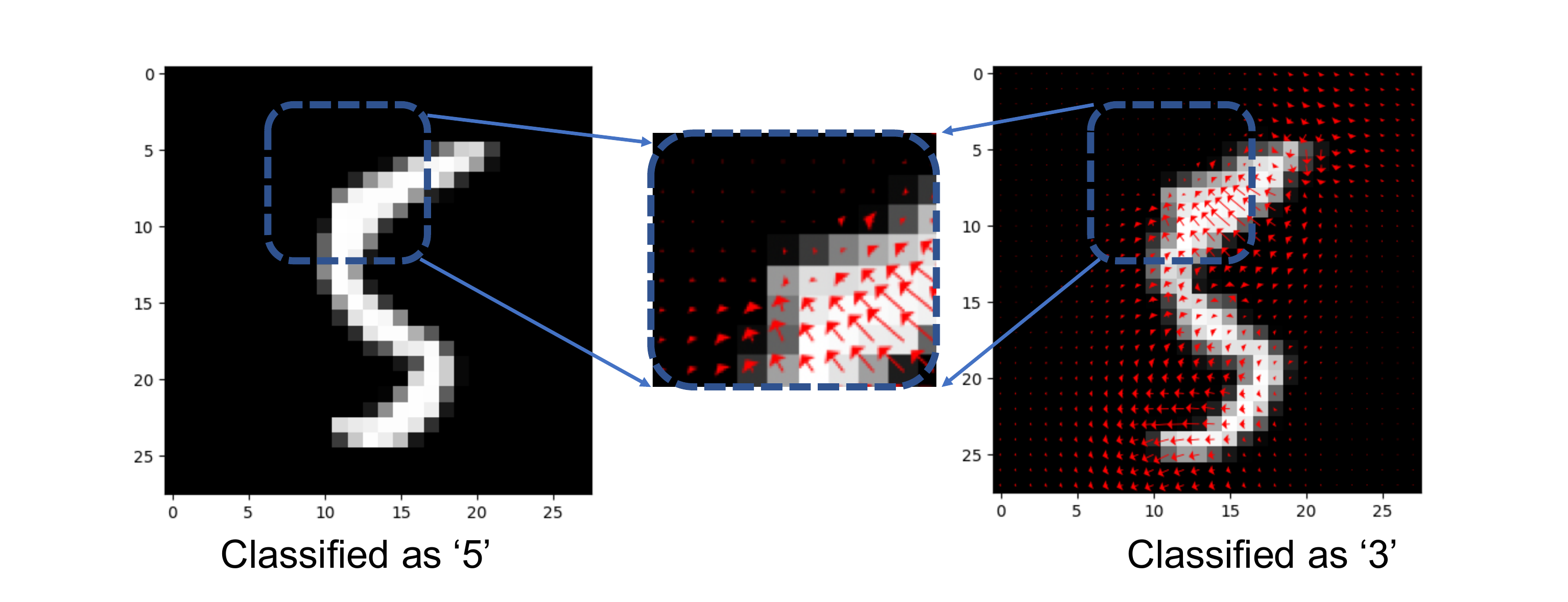}
  \caption{The top part of digit ``5"' is perturbed to be ``thicker". For the image which was correctly classified as ``5", after distortion is now classified as ``3".}
  \label{fig:spatial}
\end{figure}

Traditional adversarial attack algorithms directly modify the pixel value of an image, which changes the image's color intensity. The work~\citep{xiao2018spatially} devises another method, called a Spatially Transformed Attack. They perturb the image by doing slight spatial transformation: they translate, rotate and distort the local image features slightly. The perturbation is small enough to evade human inspection but it can fool the classifiers. One example is in figure (\ref{fig:spatial}).

\subsubsection{Unrestricted Adversarial Examples}

Previous attack methods only consider adding unnoticeable perturbations into images. However, the work~\citep{song2018constructing} introduces a method to generate unrestricted adversarial examples. These samples do not necessarily look exactly the same as the victim samples, but are still legitimate samples for human eyes and can fool the classifier. Previous successful defense strategies that target perturbation-based attacks fail to recognize them.

In order to attack a given classifier $C$, Song et al. pretrained an Auxiliary Classifier Generative Adversarial Network (AC-GAN), \citep{odena2017conditional}, so they can generate one legitimate sample $x$ from a noise vector $z^0$ from class $y$. Then, to craft an adversarial example, they will find a noise vector $z$ near $z^0$, but require that the output of AC-GAN generator $\mathcal{G}(z)$ be wrongly classified by victim model $C$. Because $z$ is near $z^0$ in the latent space of the AC-GAN, its output should belong to the same class $y$. In this way, the generated sample $\mathcal{G}(z)$ is different from $x$, misleading classifier $F$, but it is still a legitimate sample.

\subsection{Physical World Attack}
All the previously introduced attack methods are applied digitally, where the adversary supplies input images directly to the machine learning model. However, this is not always the case for some scenarios, like those that use cameras, microphones or other sensors to receive the signals as input. In this case, can we still attack these systems by generating physical-world adversarial objects? Recent works show that such attacks do exist. For example, in~\citep{eykholt2017robust}, stickers are attached to road signs that can severely threaten autonomous cars' sign recognizer. These kinds of adversarial objects are more destructive for deep learning models because they can directly challenge many practical applications of DNNs, such as face recognition, autonomous vehicle, etc.

\subsubsection{Exploring Adversarial Examples in Physical World}

In~\citep{kurakin2016adversarial1}, authors explore the feasibility of crafting physical adversarial objects, by checking whether the generated adversarial images (FGSM, BIM) are ``robust'' under natural transformation (such as changing viewpoint, lighting etc.). Here, ``robust'' means the crafted images remain adversarial after the transformation. To apply the transformation, they print out the crafted images, and let test subjects use cellphones to take photos of these printouts. In this process, the shooting angle or lighting environment are not constrained, so the acquired photos are transformed samples from previously generated adversarial examples. The experimental results demonstrate that after transformation, a large portion of these adversarial examples, especially those generated by FGSM, remain adversarial to the classifier. These results suggest the possibility of physical adversarial objects which can fool the sensor under different environments.

\subsubsection{Eykholt's Attack on Road Signs}
\begin{figure} [h]
\centering
\includegraphics[width=0.8\linewidth]{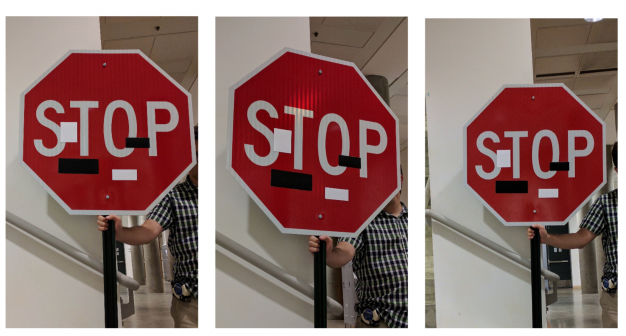}
\caption{The attacker puts some stickers on a road sign to confuse an autonomous vehicle's road sign recognizer from any viewpoint. (Image Credit: \cite{eykholt2017robust})}
\label{fig:roadsign}
\end{figure}

The work \citep{eykholt2017robust}, shown in figure \ref{fig:roadsign}, crafts physical adversarial objects by ``contaminating'' road signs to mislead road sign recognizers. They achieve the attack by putting stickers on the stop sign in the desired positions. 

The approach consists of: (1) Implement $l_1$-norm based attack (those attacks that constrain $||x'-x||_1$) on digital images of road signs to roughly find the region to perturb ($l_1$ attacks render sparse perturbation, which helps to find the attack location). These regions will later be the location of stickers. (2) Concentrate on the regions found in step 1, and use an $l_2$-norm based attack to generate the color for the stickers. (3) Print out the perturbation found in steps 1 and 2, and stick them on road sign. The perturbed stop sign can confuse an autonomous vehicle from any distance and viewpoint. 

\subsubsection{Athalye's 3D Adversarial Object}
\begin{figure}[t]
    \centering
    \includegraphics[width = 80mm]{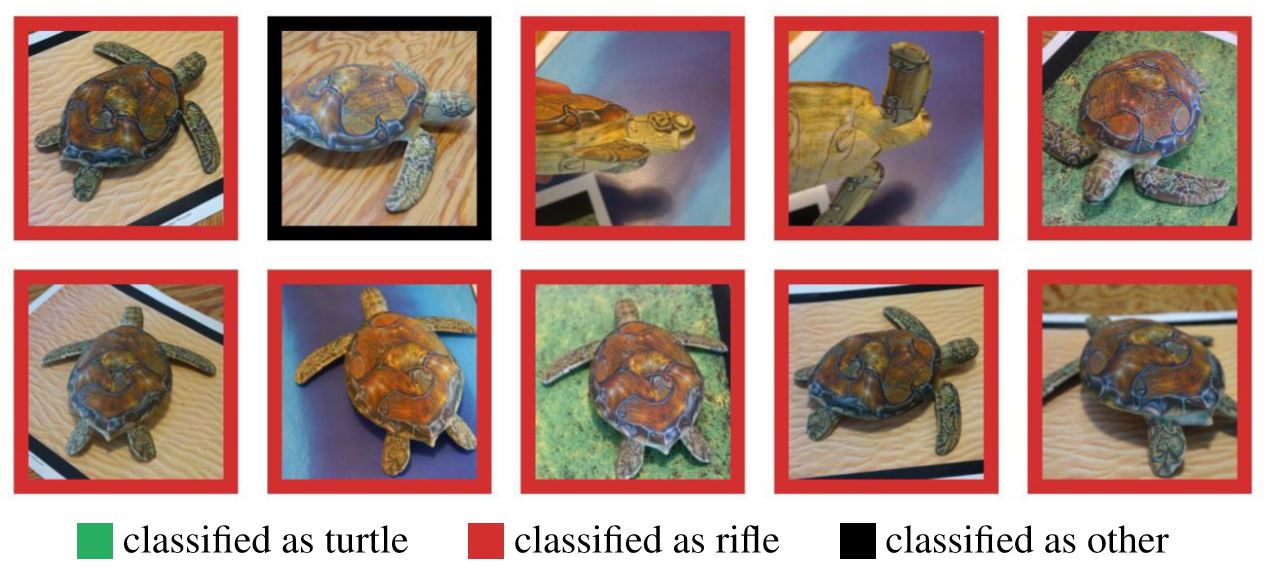}
    \caption{The image classifier fails to correctly recognize the adversarial object, but the original object can be correctly predicted with 100\% accuracy. (Image Credit: \citep{athalye2017synthesizing})}
    \label{fig:turtle}
\end{figure}

The work~\citep{athalye2017synthesizing} reported the first work which successfully crafted physical 3D adversarial objects. As shown in figure \ref{fig:turtle}, the authors use 3D-printing to manufacture an ``adversarial'' turtle. To achieve the goal, they implement a 3D rendering technique. Given a textured 3D object, they first optimize the object's texture such that the rendering images are adversarial from any viewpoint. In this process, they also ensure that the perturbation remains adversarial under different environments: camera distance, lighting conditions, rotation and background. After finding the perturbation on 3D rendering, they print an instance of the 3D object.

\subsection{Black-Box Attacks}
\subsubsection{Substitute Model}\label{Sub}
The work of \citep{papernot2017practical} was the first to introduce an effective algorithm to attack DNN classifiers, under the condition that the adversary has no access to the classifier's parameters or training set (black-box). An adversary can only feed input $x$ to obtain the output label $y$ from the classifier. Additionally, the adversary may have only partial knowledge about: (a) the classifier's data domain (e.g. handwritten digits, photographs, human faces) and (b) the architecture of the classifier (e.g., CNNs, RNNs).

The work~\citet{papernot2017practical} exploits the ``transferability'' (section \ref{transfer}) property of adversarial examples: a sample $x'$ can attack $F_1$, it is also likely to attack $F_2$, which has similar structures to $F_1$. Thus, the authors introduce a method to train a substitute model $F'$ to imitate the target victim classifier $F$, and then craft the adversarial example by attacking substitute model $F'$.  The main steps are as follows:

\begin{enumerate}[topsep=0pt,itemsep = 0pt]
    \item Synthesize Substitute Training Dataset\\
    Make a ``replica'' training set. For example, to attack a victim classifier for the hand-written digits recognition task, make an initial substitute training set by: (a) requiring samples from test set; or (b) handcrafting samples.
    \item Train the Substitute Model\\
    Feed the substitute training dataset $X$ into the victim classifier to obtain their labels $Y$. Choose one substitute DNN model to train on $(X,Y)$ to get $F'$. Based on the attacker's knowledge, the chosen DNN should have similar structures to the victim model.
    \item Augment Dataset \\
    Augment the dataset $(X,Y)$ and retrain the substitute model $F'$ iteratively. This procedure helps to increase the diversity of the replica training set and improve the accuracy of substitute model $F'$.
    \item Attack the Substitute Model\\
    Utilize the previously introduced attack methods, such as FGSM, to attack the model $F'$. The generated adversarial examples are also very likely to mislead the target model $F$, by the property of ``transferability''. 
\end{enumerate}
\textit{What kind of attack algorithm should we choose to attack substitute model?} The success of the substitute model black-box attack is based on the ``transferability'' property of adversarial examples. Thus, during black-box attack, we choose attacks that have high transferability, like FGSM, PGD and momentum-based iterative attacks \citep{dong2018boosting}. 

\subsubsection{ZOO: Zeroth Order Optimization Based Black-Box Attack}
Different from the work in section \ref{Sub} where an adversary can only obtain the label information from the classifier, it assumes that the attacker has access to the prediction confidence (score) from the victim classifier's output\citep{chen2017zoo}. In this case, there is no need to build the substitute training set and substitute model. Chen et al. give an algorithm to ``scrape'' the gradient information around victim sample $x$ by observing the changes in the prediction confidence $F(x)$ as the pixel values of $x$ are tuned.

The equation \ref{eq:zoo} shows for each index $i$ of sample $x$, we add (or subtract) $x_i$ by $h$. If $h$ is small enough, we can scrape the gradient information from the output of $F(\cdot)$ by:
\begin{equation}\label{eq:zoo}
        \pdv{F(x)}{x_i}\approx \frac{F(x+he_i)-F(x-he_i)}{2h}
\end{equation}
Utilizing the approximate gradient, we can apply the attack formulations introduced in Section \ref{fgsm} and Section \ref{cw}. The success rate of \textit{ZOO} is higher than \textit{substitute model} (Section \ref{Sub}) because it can utilize the information of prediction confidence, instead of solely the predicted labels.

\subsubsection{Query-Efficient Black-Box Attack}
Previously introduced black-box attacks require lots of input queries to the classifier, which may be prohibitive in practical applications. There are some studies on improving the efficiency of generating black-box adversarial examples via a limited number of queries. For example, a more efficient way is introduced to estimate the gradient information from model outputs \citep{ilyas2018black}. It uses Natural Evolutional Strategies \citep{wierstra2014natural} to sample the model's output based on the queries around $x$, and estimate the expectation of gradient of $F$ on $x$. This procedure requires fewer queries to the model. Moreover, a genetic algorithm is applied to search the neighbors of a benign image for adversarial examples~\citep{alzantot2018genattack}.

\subsection{Semi-white (Grey) box Attack}

In~\citep{xiao2018generating}, a semi-white box attack framework is introduced. It first trained a Generative Adversarial Network (GAN) \citep{goodfellow2014generative}, targeting the model of interest. The attacker can then craft adversarial examples directly from the generative network. The advantage of the GAN-based attack is that it accelerates the process of producing adversarial examples, and makes more natural and more undetectable samples. Later, in \citep{deb2019advfaces}, GAN is used to generate adversarial faces to evade the face recognition software. Their crafted face images appear to be more natural and have barely distinguishable difference from target face images.

\subsection{Poisoning attacks}
The attacks we have discussed so far are evasion attacks, which are launched after the classification model is trained. Some works instead craft adversarial examples before training. These adversarial examples are inserted into the training set in order to undermine the overall accuracy of the learned classifier, or influence its prediction on certain test examples. This process is called a poisoning attack. 

Usually, the adversary in a poisoning attack setting has knowledge about the architecture of the model which is later trained on the poisoned dataset. Poisoning attacks frequently applied to attack graph neural network, because of the GNN's specific transductive learning procedure. Here, we introduce studies that craft image poisoning attacks.

\subsubsection{Biggio's Poisoning Attack on SVM}
The work of \citep{biggio2012poisoning} introduced a method to poison the training set in order to reduce SVM model's accuracy. In their setting, they try to figure out a poison sample $x_c$ which, when inserted into the training set, will result in the learned SVM model $F_{x_c}$ having a large total loss on the whole validation set. They achieve this by using incremental learning techniques for SVMs \citep{cauwenberghs2001incremental}, which can model the influence of training samples on the learned SVM model. 

A poisoning attack based on procedure above is quite successful for SVM models. However, for deep learning models, it is not easy to explicitly figure out the influence of training samples on the trained model. Below we introduce some approaches for applying poisoning attacks on DNN models.

\subsubsection{Koh's Model Explanation}

In~\citep{koh2017understanding}, a method is introduced to interpret deep neural networks: how would the model's predictions change if a training sample were modified? Their model can explicitly quantify the change in the final loss without retraining the model when only one training sample is modified. This work can be naturally adopted to poisoning attacks by finding those training samples that have large influence on model's prediction. 

\subsubsection{Poison Frogs}
In~\citep{shafahi2018poison}, a method is introduced to insert an adversarial image with true label to the training set, in order to cause the trained model to wrongly classify a target test sample. In this work, given a target test sample $x_t$ with the true label $y_t$, the attacker first uses a base sample $x_b$ from class $y_b$. Then, it solves the objective to find $x'$:
\begin{equation*}
    x' = \argmin_x ||Z(x)-Z(x_t)||_2^2+\beta ||x-x_b||_2^2. 
\end{equation*}
After inserting the poison sample $x'$ into the training set, the new model trained on $X_{train}+\{x\}'$ will classify $x'$ as class $y_b$, because of the small distance between $x'$ and $x_b$. Using a new trained model to predict $x_t$, the objective of $x'$ forces the score vector of $x_t$ and $x'$ to be close. Thus, $x'$ and $x_t$ will have the same prediction outcome. In this way, the new trained model will predict the target sample $x_t$ as class $y_b$.

\section{Countermeasures Against Adversarial Examples} \label{defend}

In order to protect the security of deep learning models, different strategies have been considered as countermeasures against adversarial examples. There are basically three main categories of these countermeasures:
\begin{enumerate}
    \item Gradient Masking/Obfuscation\\
    Since most attack algorithms are based on the gradient information of the classifier, masking or hiding the gradients will confound the adversaries. 
    \item Robust Optimization\\
    Re-learning a DNN classifier's parameters can increase its robustness. The trained classifier will correctly classify the subsequently generated adversarial examples.
    \item Adversarial Examples Detection \\
    It studies the distribution of natural/benign examples, detects adversarial examples and disallows their input into the classifier
\end{enumerate}

\subsection{Gradient Masking/Obfuscation}\label{mask}
\textit{Gradient masking/obfuscation} refers to the strategy where a defender deliberately hides the gradient information of the model, in order to confuse the adversaries, since most attack algorithms are based on the classifier's gradient information.

\subsubsection{Defensive Distillation}\label{dd}
``Distillation'', first introduced by \citep{hinton2015distilling}, is a training technique to reduce the size of DNN architectures. It fulfills its goal by training a smaller-size DNN model on the logits (outputs of the last layer before softmax). 

In~\citep{papernot2016distillation}, it reformulates the procedure of distillation to train a DNN model that can resist adversarial examples, such as FGSM, Szegedy's L-BFGS Attack or DeepFool. It designs the training process as:
 \begin{enumerate}[itemsep=0mm]
        \item [(1)] Train a network $F$ on the given training set $(X,Y)$ by setting the temperature\footnote{Note that the softmax function at a temperature $T$ means:
        $softmax  (x,T)_i = \frac{e^{x_i/T}}{\sum_{j} e^{x_j/T}}$, where $i=0,2,...,K-1$} of the softmax to $T$.
        \item [(2)] Compute the scores (after softmax) given by  $F(X)$ again and  evaluate the scores at temperature $T$.
        \item [(3)] Train another network $F'_T$ using softmax at temperature $T$ on the dataset with soft labels $(X,F(X))$. We refer the model $F'_T$ as the \textit{distilled model}.
        \item [(4)] Use the distilled network $F'_T$ with softmax at temperature 1, which is denoted as $F'_1$ during prediction on test data $X_{test}$ (or adversarial examples),
    \end{enumerate}
    In~\citet{carlini2017towards}, it explains why this algorithm works: When we train a distilled network $F'_T$ at temperature T and test it at temperature 1, we effectively cause the inputs to the softmax to become larger by a factor of T. Let us say $T=100$, the logits $Z(\cdot)$ for sample $x$ and its neighbor points $x'$ will be 100 times larger. It will lead to the softmax function $F_1(\cdot) = \text{softmax}(Z(\cdot),1)$ outputting a score vector like $(\epsilon, \epsilon, ..., 1-(m-1)\epsilon, \epsilon,...,\epsilon)$, where the target output class has a score extremely close to 1, and all other classes have scores close to 0. In practice, the value of $\epsilon$ is so small that its 32-bit floating-point value for computer is rounded to 0. In this way, the computer cannot find the gradient of score function $F'_1$, which inhibits the gradient-based attacks.

\subsubsection{Shattered Gradients}\label{sg}
Some studies\citep{buckman2018thermometer,guo2017countering} try to protect the model by preprocessing the input data. They add a non-smooth or non-differentiable preprocessor $g(\cdot)$ and then train a DNN model $f$ on $g(X)$. The trained classifier $f(g(\cdot))$ is not differentiable in term of $x$, causing the failure of adversarial attacks. 

For example, \textit{Thermometer Encoding} \citep{buckman2018thermometer} uses a preprocessor to discretize an image's pixel value $x_i$ into a $l$-dimensional vector $\tau(x_i)$. (e.g. when $l = 10$, $\tau (0.66) = 1111110000$). The vector $\tau(x_i)$ acts as a ``thermometer'' to record the pixel $x_i$'s value. A DNN model is later trained on these vectors. The work in~\citep{guo2017countering} studies a number of image processing tools, such as image cropping, compressing or total-variance minimization, to determine whether these techniques help to protect the model against adversarial examples. All these approaches block up the smooth connection between the model's output and the original input samples, so the attacker cannot easily find the gradient $\pdv{F(x)}{x}$ for attacking.

\subsubsection{Stochastic/Randomized Gradients} \label{srg}
Some defense strategies try to randomize the DNN model in order to confound the adversary. For instance, we train a set of classifiers $s = \{F_t:t=1,2,...,k\}$. During evaluation on data $x$, we randomly select one classifier from the set $s$ and predict the label $y$. Because the adversary has no idea which classifier is used by the prediction model, the attack success rate will be reduced. 

Some examples of this strategy include the work in~\citep{dhillon2018stochastic} that randomly drops some neurons of each layer of the DNN model, and the work in~\citep{xie2017mitigating} that resizes the input images to a random size and pads zeros around the input image.

\subsubsection{Exploding \& Vanishing Gradients}\label{evg}
Both \textit{PixelDefend} \citep{song2017pixeldefend} and \textit{Defense-GAN} \citep{samangouei2018defense} suggest using generative models to project a potential adversarial example onto the benign data manifold before classifying them. While \textit{PixelDefend} uses PixelCNN generative model \citep{van2016conditional}, \textit{Defense-GAN} utilizes a GAN architecture \citep{goodfellow2014generative}. The generative models can be viewed as a purifier that transforms adversarial examples into benign examples.

Both of these methods add a generative network before the classifier DNN, which will cause the final classification model to be an extremely deep neural network. The underlying reason that these defenses succeed is because: the cumulative product of partial derivatives from each layer
will cause the gradient $\pdv{\mathcal{L}(x)}{x}$ to be extremely small or irregularly large, which prevents the attacker accurately estimating the location of adversarial examples. 

\subsubsection{Gradient Masking/Obfuscation Methods are not Safe}
The work in~\citep{carlini2017towards} shows that the method of ``Defensive Distillation'' (section \ref{dd}) is still vulnerable to their adversarial examples. The work in~\citep{athalye2018obfuscated} devises different attacking algorithms to break gradient masking/obfuscation defending strategies (Section (\ref{sg}), Section (\ref{srg}), Section (\ref{evg})). The main weakness of the gradient masking strategy is that: it can only ``confound'' the adversaries, but it cannot eliminate the existence of adversarial examples.
\subsection{Robust Optimization} \label{training}

Robust optimization methods aim to improve the classifier's robustness (Section (\ref{rob})) by changing DNN model's manner of learning. They study how to learn model parameters that can give promising predictions on potential adversarial examples. In this field, the works majorly focus on: (1) learning model parameters $\theta^*$ to minimize the average \text{adversarial loss}: (Section \ref{adr})
\begin{equation}\label{eq:advloss}
    \theta^* = \argmin_{\theta\in\Theta}  \displaystyle \mathop{\mathbb{E}}_{x\sim \mathcal{D}}~ \max_{||x'-x||\leq\epsilon}~\mathcal{L}  (\theta,x',y)
\end{equation}

or (2) learning model parameters $\theta^*$ to maximize the average \textit{minimal perturbation distance}: (Section \ref{mip})
\begin{equation}\label{eq:mp}
    \theta^* = \argmax_{\theta\in \Theta}  \displaystyle \mathop{\mathbb{E}}_{x\sim \mathcal{D}}~ \min_{C(x')\neq y}~||x'-x||.
\end{equation}

Typically, a robust optimization algorithm should have a prior knowledge of its potential threat or potential attack (adversarial space $\mathcal{D}$). Then, the defenders build classifiers which are safe against this specific attack. For most of the related works \citep{goodfellow2014explaining,kurakin2016adversarial1,madry2017towards}, they aim to defend against adversarial examples generated from small $l_p$ (specifically $l_\infty$ and $l_2$) norm perturbation. Even though there is a chance that these defenses are still vulnerable to attacks from other mechanisms, (e.g. \citep{xiao2018spatially}), studying the security against $l_p$ attack is fundamental and can be generalized to other attacks. 

In this section, we concentrate on defense approaches using robustness optimization against $l_p$ attacks. We categorize the related works into three groups: (a) regularization methods, (b) adversarial (re)training and (c) certified defenses.

\subsubsection{Regularization Methods}
Some early studies on defending against adversarial examples focus on exploiting certain properties that a robust DNN should have in order to resist adversarial examples. For example, in~\citep{szegedy2013intriguing}, it suggests that a robust model should be stable when its inputs are distorted, so they turn to constrain the \textit{Lipschitz constant} to impose this ``stability'' of model output. Training on these regularizations can sometimes heuristically help the model be more robust. 

\begin{enumerate}
    \item Penalize Layers' Lipschitz Constant\\
    When first claimed the vulnerability of DNN models to adversarial examples, authors in \citep{szegedy2013intriguing} suggested that adding regularization terms on the parameters during training can force the trained model to be stable. It suggested constraining the Lipschitz constant $L_k$ between any two layers: 
    \begin{equation*}
    \forall x,\delta,~~~||h_k(x;W_k)-h_k(x+\delta;W_k)||
    \leq L_k ||\delta||,
    \end{equation*}
    so that the outcome of each layer will not be easily influenced by the small distortion of its input. The work \citep{cisse2017parseval} formalized this idea, by claiming that the model's \textit{adversarial risk} (\ref{eq:advloss}) is right dependent on this instability $L_k$:
    \begin{equation*}
    \begin{split}
        &\displaystyle \mathop{\mathbb{E}}_{x\sim \mathcal{D}}~\mathcal{L}_{adv}(x) \leq \displaystyle \mathop{\mathbb{E}}_{x\sim \mathcal{D}}~\mathcal{L}(x)\\
        & + \displaystyle \mathop{\mathbb{E}}_{x\sim \mathcal{D}} [\max_{||x'-x||\leq\epsilon}|\mathcal{L}(F(x'),y) - \mathcal{L}(F(x),y)|]\\
        & \leq\displaystyle \mathop{\mathbb{E}}_{x\sim \mathcal{D}}~\mathcal{L}(x) + \lambda_p\prod_{k=1}^{K} L_k
    \end{split}
    \end{equation*}
    where $\lambda_p$ is the Lipschitz constant of the loss function. This formula states that during the training process, penalizing the large instability for each hidden layer can help to decrease the adversarial risk of the model, and consequently increase the robustness of model. The idea of constraining instability also appears in the work \citep{miyato2015distributional} for semi-supervised, and unsupervised defenses.
    
    \item Penalize Layers' Partial Derivative\\
    The work in~\citep{gu2014towards} introduced a Deep Contractive Network algorithm to regularize the training. It was inspired by the Contractive Autoencoder \citep{rifai2011contractive}, which was introduced to denoise the encoded representation learning. The Deep Contractive Network suggests adding a penalty on the partial derivatives at each layer into the standard back-propagation framework, so that the change of the input data will not cause large change on the output of each layer. Thus, it becomes difficult for the classifier to give different predictions on perturbed data samples.

\end{enumerate}

\subsubsection{Adversarial (Re)Training}
\begin{enumerate}[label=(\roman*)]
\item Adversarial Training with FGSM\label{advtf1}\\

The work in ~\citep{goodfellow2014explaining} is the first to suggest feeding generated adversarial examples into the training process. By adding the adversarial examples with true label $(x',y)$ into the training set, the training set will tell the classifier that $x'$ belongs to class $y$, so that the trained model will correctly predict the label of future adversarial examples.

In the work \citep{goodfellow2014explaining}, they use non-targeted FGSM (Section (\ref{fgsm})) to generate adversarial examples $x'$ for the training dataset:
\begin{equation*} 
 x' = x + \epsilon \text{sign}(\nabla_x \mathcal{L}(\theta,x,y)),
\end{equation*}
By training on benign samples augmented with adversarial examples, they increase the robustness against adversarial examples generated by FGSM.

The training strategy of this method is changed in~\citep{kurakin2016adversarial1} so that the model can be scaled to larger dataset such as ImageNet. They suggest that using batch normalization \citep{ioffe2015batch} should improve the efficiency of adversarial training. We give a short sketch of their algorithm in Algorithm (\ref{advtraining}).
\begin{algorithm}[t]
\caption{Adversarial Training with FGSM by batches}
\label{advtraining}
\begin{algorithmic}
\STATE Randomly initialize network $F$
\REPEAT 
\STATE 1. Read minibatch $B = \{x^1,...,x^m\}$ from training set
\STATE 2. Generate $k$ adversarial examples $\{x_{adv}^1,...x_{adv}^k\}$ for corresponding benign examples using current state of the network $F$.
\STATE 3. Update  $B' = \{x_{adv}^1,...,x_{adv}^k,x^{k+1},...,x^m\}$
\STATE Do one training step of network $F$ using minibatch $B'$
\UNTIL{training converged}
\end{algorithmic}
\end{algorithm}

The trained classifier has good robustness on FGSM attacks, but it is still vulnerable to iterative attacks. Later, the work in~\citep{tramer2017ensemble} argues that this defense is also vulnerable to single-step attacks. Adversarial training with FGSM will cause gradient obfuscation (Section (\ref{mask})), where there is an extreme non-smoothness of the trained classifier $F$ near the test sample $x$.   

\begin{figure} [b] 
  \centering
  \includegraphics[width=\linewidth]{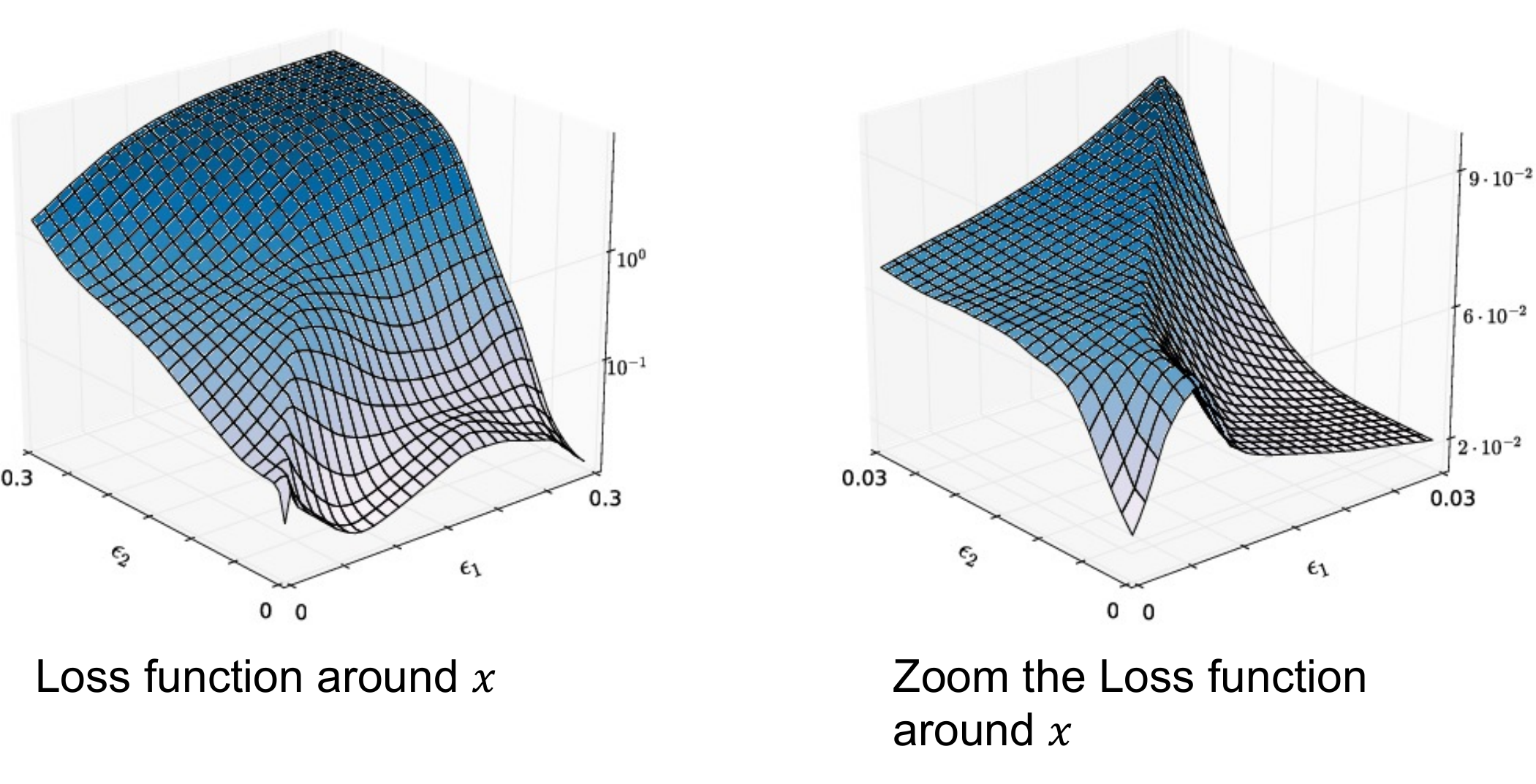}
  \caption{Illustration of gradient masking for adversarial training via FGSM. It plots the loss function of the trained classifier around $x$ on the grids of gradient direction and another randomly chosen direction. We can see that the gradient poorly approximates the global loss. (Image Credit: \citep{tramer2017ensemble})}
  \label{fig:mask}
\end{figure}

\item Adversarial Training with PGD \label{advtp}

The work in~\citep{madry2017towards} suggests using Projected Gradient Descent attack (Section (\ref{bim})) for adversarial training, instead of using single-step attacks like FGSM. The PGD attacks (Section (\ref{bim})) can be seen as a heuristic method to find the ``most adversarial'' example in the $l_\infty$ ball around x: $B_{\epsilon}(x)$: 
\begin{equation}\label{eq:mostadv}
     x_{adv} = \argmax_{x'\in B_{\epsilon}(x)}\mathcal{L}(x',F)
\end{equation}
Here, the most-adversarial example $x_{adv}$ is the location where the classifier $F$ is most likely to be misled.
When training the DNN model on these most-adversarial examples, it actually solves the problem of learning model parameters $\theta$ that minimize the adversarial loss (\ref{eq:advloss}). If the trained model has small loss value on these most-adversarial examples, the model is safe at everywhere in $x$'s neighbor ball $B_{\epsilon}(x)$. One thing to note is: this method trains the model only on adversarial examples, instead of a mix of benign and adversarial examples. The training algorithm is shown in Algorithm (\ref{madry}).

\begin{algorithm}[b]
\caption{Adversarial Training with  PGD}
\label{madry}
\begin{algorithmic}
\STATE Randomly initialize network $F$
\REPEAT 
\STATE 1. Read minibatch $B = \{x^1,...,x^m\}$ from training set
\STATE 2. Generate $m$ adversarial examples $\{x_{adv}^1,...x_{adv}^m\}$ by PGD attack using current state of the network $F$.
\STATE 3. Update  $B' = \{x_{adv}^1,...,x_{adv}^m\}$
\STATE Do one training step of network $F$ using minibatch $B'$
\UNTIL{training converged}
\end{algorithmic}
\end{algorithm}

The trained model under this method demonstrates good robustness against both single-step and iterative attacks on MNIST and CIFAR10 datasets. However, this method involves an iterative attack for all the training samples. Thus, the time complexity of this adversarial training will be $k$ (using $k$-step PGD) times as large as that for natural training, and as a consequence, it is hard to scale to large datasets such as ImageNet. 

\item Ensemble Adversarial Training \label{ens}

The work in~\citep{tramer2017ensemble} introduced an adversarial training method which can protect CNN models against single-step attacks and can be also applied to large datasets such as ImageNet. 

The main approach is to augment the classifier's training set with adversarial examples crafted from other pre-trained classifiers. For example, if we aim to train a robust classifier $F$, we can first pre-train classifiers $F_1$, $F_2$, and $F_3$ as references. These models have different hyperparameters with model $F$. Then, for each sample $x$, we use a single-step attack FGSM to craft adversarial examples on $F_1$, $F_2$ and $F_3$ to get $x_{adv}^1$, $x_{adv}^2$, $x_{adv}^3$. Because of the transferability property (section \ref{transfer}) of the single-step attacks across different models, $x_{adv}^1$, $x_{adv}^2$, $x_{adv}^3$ are also likely to mislead the classifier $F$. It means that these samples are a good approximation for the ``most adversarial'' example (\ref{eq:mostadv}) for model $F$ on $x$. Training on these samples together will approximately minimize the adversarial loss in (\ref{eq:advloss}).

This ensemble adversarial training algorithm is more efficient than these in Section \ref{advtf1} and Section \ref{advtp}, since it decouples the process of model training and generating adversarial examples. The experimental results show that this method can provide robustness against single-step and black-box attacks on the ImageNet dataset.

\item Accelerate Adversarial Training \label{free}

While it is one of the most promising and reliable defense strategies, adversarial training with PGD attack \citep{madry2017towards} is generally slow and computationally costly.

The work in~\citep{shafahi2019adversarial} proposes a free adversarial training algorithm which improves the efficiency by reusing the backward pass calculations. In this algorithm, the gradient of the loss to input: $\pdv{\mathcal{L}(x+\delta,\theta)}{x}$ and the gradient of the loss to model parameters: $\pdv{\mathcal{L}(x+\delta,\theta)}{\theta}$ can be computed together in one back propagation iteration, by sharing the same components of chain rule. Thus, the adversarial training process is highly accelerated. The free adversarial training algorithm is shown in Algorithm (\ref{freea}).

The work in~\citep{zhang2019you} argues that when the model parameters are fixed, the PGD-generated adversarial example is only coupled with the weights of the first layer of DNN. It is based on solving a Pontryagin's Maximal Principle \citep{pontryagin2018mathematical}.
Therefore, the work in~\citep{zhang2019you} develops an algorithm You Only Propagate Once (YOPO) to reuse the gradient of the loss to the model's first layer output $\pdv{\mathcal{L}(x+\delta,\theta)}{Z_1(x)}$ during generating PGD attacks. In this way, YOPO avoids a large amount of times to access the gradient and therefore it reduces the computational cost.

\end{enumerate}

\begin{algorithm}[t]
\caption{Free Adversarial Training}
\label{freea}
\begin{algorithmic}
\STATE Randomly initialize network $F$
\REPEAT 
\STATE 1. Read minibatch $B = \{x^1,...,x^m\}$ from training set
\STATE 2. for $i=1...m$ do
\STATE $~~~~~~~~$ (a) Update model parameter $\theta$
\STATE $~~~~~~~~~~~~~~~~~$  $g_{\theta} \leftarrow \mathbb{E}_{(x,y)\in B}[\nabla_{\theta}\mathcal{L}(x+\delta, y,\theta)]$
\STATE $~~~~~~~~~~~~~~~~~$  $g_{adv} \leftarrow \nabla_{x}\mathcal{L}(x+\delta, y,\theta)$
\STATE $~~~~~~~~~~~~~~~~~$  $\theta\leftarrow \theta - \alpha g_{\theta}$
\STATE $~~~~~~~~$ (b) Generate adversarial examples
\STATE $~~~~~~~~~~~~~~~~~$  $\delta \leftarrow \delta+\epsilon \cdot sign(g_{adv})$
\STATE $~~~~~~~~~~~~~~~~~$  $\delta \leftarrow clip(\delta, -\epsilon, \epsilon)$
\STATE 3. Update minibatch $B$ with adversarial examples $x+\delta$
\UNTIL{training converged}
\end{algorithmic}
\end{algorithm}

\subsubsection{Provable Defenses}
Adversarial training has been shown to be effective in protecting models against adversarial examples. However, this is still no formal guarantee about the safety of the trained classifiers. We will never know whether there are more aggressive attacks that can break those defenses, so directly applying these adversarial training algorithms in safety-critical tasks would be irresponsible. 

As we mentioned in Section (\ref{ground}), the work \citep{carlini2017provably} was the first to introduce a Reluplex algorithm to seriously verify the robustness of DNN models: when the model $F$ is given, the algorithm figures out the exact value of minimal perturbation distance $r(x;F)$. This is to say, the classifier is safe against any perturbations with norm less than $r(x;F)$. If we apply Reluplex on the whole test set, we can tell what percentage of samples are absolutely safe against perturbations less than norm $r_0$. In this way, we gain confidence and reduce the expected risk when building DNN models.

The method of Reluplex seeks to find the exact value of $r(x;F)$ that can verify the model $F$'s robustness on $x$. Alternately, works such as \citep{raghunathan2018certified,wong2017provable,hein2017formal} try to find trainable ``certificates'' $\mathcal{C}(x;F)$ to verify the model robustness. For example, in \citep{hein2017formal}, a certificate $\mathcal{C}(x,F)$ is calculated for model $F$ on $x$, which is a lower bound of minimal perturbation distance: $\mathcal{C}(x,F)\leq r(x,F)$. As shown in Figure (\ref{fig:formal}), the model must be safe against any perturbation with norm limited by $\mathcal{C}(x,F)$. Moreover, these certificates are trainable. Training to optimize these certificates will grant good robustness to the classifier. In this section, we'll briefly introduce some methods to design these certificates.

\begin{enumerate}[label=(\roman*)]
    \item Lower Bound of Minimal Perturbation\\
    The work in \citep{hein2017formal} derives a lower bound $\mathcal{C}(x,F)$ for the minimal perturbation distance of $F$ on $x$ based on \textit{Cross-Lipschitz Theorem}: 
    \small
    \begin{equation*}
    \max_{\epsilon>0} \, \min\{\min_{i \neq y}\frac {Z_y(x)-Z_i(x)}{ \max_{x'\in B_{\epsilon}(x)}||\nabla Z_y(x')-\nabla Z_i(x')||},\, \epsilon \}       
    \end{equation*}
    \normalsize
    The detailed derivation can be found in \citep{hein2017formal}. Note that the formulation of $\mathcal{C}(x,F)$ only depends on $F$ and $x$, and it is easy to calculate for a neural network with one hidden layer. The model $F$ thus can be proved to be safe in the region within distance $\mathcal{C}(x,F)$. Training to maximize this lower bound will make the classifier more robust.
    
\begin{figure} [t] 
  \centering
  \includegraphics[width=0.7\linewidth]{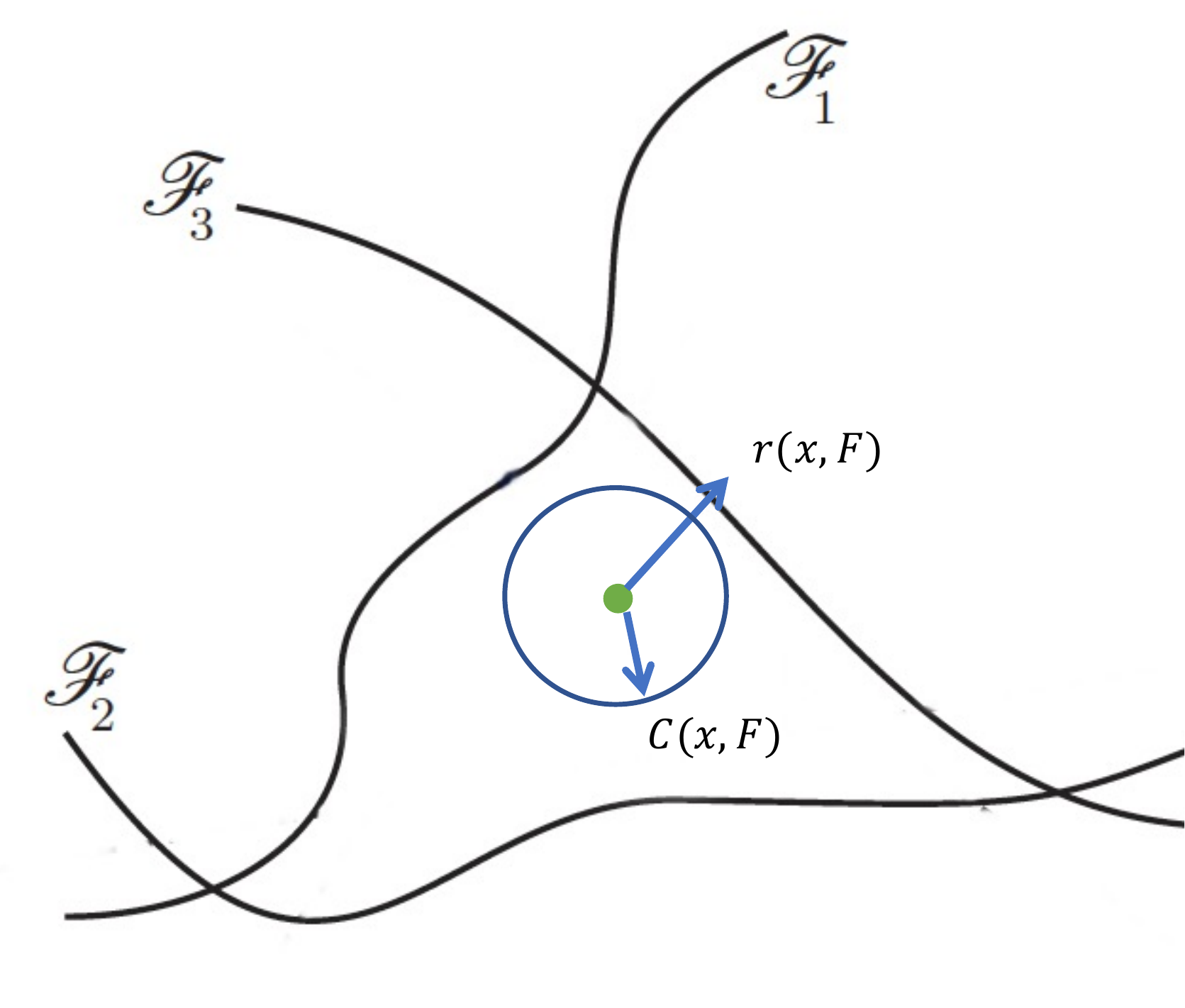}
  \caption{The derived certificate $\mathcal{C}(x,F)$ is a lower bound of minimal perturbation distance $\rho(x,F)$. The model is safe in $\mathcal{C}(x,F)$ ball. }
  \label{fig:formal}
\end{figure}       

    \item Upper Bound of Adversarial Loss \\
    The works in~\citep{raghunathan2018certified,wong2017provable} aim to solve the same problem. They try to find an upper bound $\mathcal{U}(x,F)$ which is larger than adversarial loss $\mathcal{L}_{adv}(x,F)$:
    \begin{equation}\label{eq:upper}
    \begin{split}
    & \mathcal{L}_{adv}(x) = \max_{x'} \, \{\max_{i\neq y} Z_i(x')-Z_y(x')\}\\
    & \text{subject to} ~~~ x' \in B_{\epsilon}(x)
    \end{split}
    \end{equation}
    Recall that in Section (\ref{adr}), we introduced the function $\max_{i\neq y} Z_i(x')-Z_y(x')$ as a type of loss function called \textit{margin loss}.
    
    The certificate $\mathcal{U}(x,F)$ acts in this way: if $\mathcal{U}(x,F)<0$, then adversarial loss $\mathcal{L}(x,F)<0$. Thus, the classifier always gives the largest score to the true label $y$ in the region $B_\epsilon(x)$, and the model is safe in this region. To increase the model's robustness, we should learn parameters that have the smallest $\mathcal{U}$ value, so that more and more data samples will have negative $\mathcal{U}$ values.
    
    The work \citep{raghunathan2018certified} uses integration inequalities to derive the certificate and use semi-definite programming (SDP) \citep{vandenberghe1996semidefinite} to solve the certificate. In contrast, the work \citep{wong2017provable} transforms the problem (\ref{eq:upper}) into a linear programming problem and solves the problem via training an alternative neural network. Both methods only consider neural networks with one hidden layer. There are also studies \citep{raghunathan2018semidefinite,wong2018scaling} that improved the efficiency and scalability of these algorithms.
    
    Furthermore, the work in ~\citep{sinha2017certifying} combines adversarial training and provable defense together. They train the classifier by feeding adversarial examples which are sampled from the distribution of worst-case perturbation, and derive the certificates by studying the Lagrangian duality of adversarial loss.

\end{enumerate}

\subsection{Adversarial Example Detection} \label{detection}

Adversarial example detection is another main approach to protect DNN classifiers. Instead of predicting the model's input directly, these methods first distinguish whether the input is benign or adversarial. Then, if it can detect that the input is adversarial, the DNN classifier will refuse to predict its label. In \citep{carlini2017adversarial}, it sorts the threat models into 3 categories that the detection techniques should deal with:
\begin{enumerate}
    \item \textit{A Zero-Knowledge Adversary} only has access to the classifier $F$'s parameters, and has no knowledge of the detection model $D$. 
    \item \textit{A Perfect-Knowledge Adversary} is aware of the model $F$, and the detection scheme $D$ and its parameters.
    \item \textit{A Limited-Knowledge Adversary} is aware of the model $F$ and the detection scheme $D$, but it does not have access to $D$'s parameters. That is, this adversary does not know the model's training set.
\end{enumerate}
In all these threat settings, the detection tools are required to correctly classify the adversarial examples, and have low possibility of misclassifying benign examples. Next, we will go through some main methods for adversarial example detection.

\subsubsection{An Auxiliary Model to Classify Adversarial Examples }

Some works focus on designin auxiliary models that aim to distinguish adversarial examples from benign examples. The work in~\citep{grosse2017statistical} trains a DNN model with $|\mathcal{Y}| = K+1$ labels, with an additional label for all adversarial examples, so that network will assign adversarial examples into the $K+1$ class. Similarly, the work \citep{gong2017adversarial} trains a binary classification model to discriminate all adversarial examples apart from benign samples, and then trains a classifier on recognized benign samples. The work in~\citep{metzen2017detecting} proposes a detection method to construct an auxiliary neural network $D$ which takes inputs from the values of hidden nodes $\mathcal{H}$ of the natural trained classifier. The trained detection classifier $D: \mathcal{H} \rightarrow [0,1]$ is a binary classification model that distinguishes adversarial examples from benign ones by the hidden layers. 

\subsubsection{Using Statistics to Distinguish Adversarial Examples}

Some early works heuristically study the differences in the statistical properties of adversarial examples and benign examples. For example, in~\citep{hendrycks2016early}, adversarial examples are found to place a higher weight on the larger (later) principle components where the natural images have larger weight on early principle components. Thus, they can split them by PCA. 

In \citep{grosse2017statistical}, it uses a statistical test: Maximum Mean Discrepancy (MMD) test \citep{gretton2012kernel}, which is used to test whether two datasets are drawn from the same distribution. They use this testing tool to test whether a group of data points are benign or adversarial.

\subsubsection{Checking the Prediction Consistency}

Other studies focus on checking the consistency of the sample $x$'s prediction outcome. They usually manipulate the model parameters or the input examples themselves, to check whether the outputs of the classifier have significant changes. These are based on the belief that the classifier will have stable predictions on natural examples under these manipulations.

The work in \citep{feinman2017detecting} randomizes the classifier using Dropout~\citep{srivastava2014dropout}. If these classifiers give very different prediction outcomes on $x$ after randomization, this sample $x$ is very likely to be an adversarial one. 

\begin{figure} [ht] 
  \centering
  \includegraphics[width=0.9\linewidth]{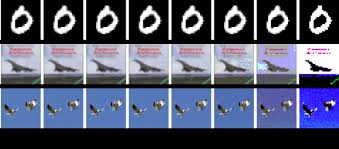}
  \caption{Images from MNIST and CIFAR10. From left to right, the color depth is reduced from 8-bit, 7-bit,...,2-bit,1-bit. (Image Credit: \citep{xu2017feature})}
  \label{fig:squeez}
\end{figure}

The work in \citep{xu2017feature} manipulates the input sample itself to check the consistency. For each input sample $x$, it reduces the color depth of the image (e.g. one 8-bit grayscale image with 256 possible values for each pixel becomes a 7-bit with 128 possible values), as shown in Figure \ref{fig:squeez}.
It hypothesizes that for natural images, reducing the color depth will not change the prediction result, but the prediction on adversarial examples will change. In this way, they can detect adversarial examples. Similar to reducing the color depth, the work \citep{feinman2017detecting} also introduced other feature squeezing methods such as spatial smoothing.

\subsubsection{Some Attacks which Evade Adversarial Detection}
The work in \citep{carlini2017adversarial} bypassed 10 of the detection methods which fall into the three categories above. The feature squeezing methods were broken by \citep{sharma2018bypassing}, which introduced a ``stronger'' adversarial attack.

The authors in \citep{carlini2017adversarial} claim that the properties which are intrinsic to adversarial examples are not very easy to find. They also gave several suggestions on future detection works:
\small
\begin{enumerate}[itemsep = 0pt]
    \item Randomization can increase the required attacking distortion.
    \item Defenses that directly manipulate on raw pixel values are ineffective.
    \item Evaluation should be down on multiple datasets besides MNIST.
    \item Report false positive and true positive rates for detection.
    \item Use a strong attack and simply focusing on white-box attacks is risky.
\end{enumerate}
\normalsize

\section{Explanations for the Existence of Adversarial Examples} \label{explain}

In addition to crafting adversarial examples and defending them, explaining the reason behind these phenomena is also important. In this section, we briefly introduce the recent works and hypotheses on the key questions of adversarial learning. We hope our introduction will give our audience a basic view on the existing ideas and solutions for these questions.

\subsection{Why Do Adversarial Examples Exist?}
Some original works such as \citep{szegedy2013intriguing}, state that the existence of adversarial examples is due to the fact that DNN models do not generalize well in low probability space of data. The generalization issue may be caused by the high complexity of DNN model structures. 

However, even linear models are also vulnerable to adversarial attacks~\citep{goodfellow2014explaining}. Furthermore, in the work \citep{madry2017towards}, they implement experiments to show that an increase in model capacity will improve the model robustness. 

Some insight can be gained about the existence of adversarial examples by studying the model's decision boundary. The adversarial examples are almost always close to decision boundary of a natural trained model, which may be because the decision boundary is too flat \citep{fawzi2016robustness}, too curved \citep{moosavi2017analysis}, or inflexible \citep{fawzi2018analysis}. 

Studying the reason behind the existence of adversarial examples is important because it can guide us in designing more robust models, and help us to understand existing deep learning models. However, there is still no consensus on this problem. 

\subsection{Can We Build an Optimal Classifier?}

Many recent works hypothesize that it might be impossible to build optimally robust classifier. For example, the work in~\citep{shafahi2018adversarial} claims that adversarial examples are inevitable because the distribution of data in each class is not well-concentrated, which leaves room for adversarial examples. In this vein, the work \citep{schmidt2018adversarially} claims that to improve the robustness of a trained model, it is necessary to collect more data. Moreover, the work in \citep{tsipras2018robustness} suggests that even if we can build models with high robustness, it must take cost of some accuracy.

\subsection{What is Transferability?}\label{transfer}
Transferability is one of the key properties of adversarial examples. It means that the adversarial examples generated to target one victim model also have a high probability of misleading other models. 

Some works compare the transferability between different attacking algorithms. In \citep{kurakin2016adversarial}, they claim that in ImageNet, single step attacks (FGSM) are more likely to transfer between models than iterative attacks (BIM) under same perturbation intensity.  

The property of transferability is frequently utilized in attacking techniques in black-box setting \citep{papernot2017practical}. If the model parameters are veiled to attackers, they can turn to attack other substitute models and enjoy the transferability of their generated samples. The property of transferability is also utilized by defending methods as in \citep{hendrycks2016early}: since the adversarial examples for model $A$ are also likely to be adversarial for model $B$, adversarial training using adversarial examples from $B$ will help defend $A$.

\section{Graph Adversarial Examples}\label{graph}

Adversarial examples also exist in graph-structured data \citep{zugner2018adversarial,dai2018adversarial}. Attackers usually slightly modify the graph structure and node features, in an effort to cause the graph neural networks (GNNs) to give wrong prediction for node classification or graph classification tasks. These adversarial attacks therefore raise concerns on the security of applying GNN models. For example, a bank needs to build a reliable credit evaluation system where their model should not be easily attacked by malicious manipulations.

There are some distinct differences between attacking graph models and attacking traditional image classifiers:
\begin{itemize}[itemsep = 0 pt]
  \item \textbf{Non-Independence} Samples of the graph-structured data are not independent: changing one's feature or connection will influence the prediction on others. 
  \item \textbf{Poisoning Attacks} Graph neural networks are usually performed in a transductive learning setting for node classification: the test data are also used to trained the classifier. This means that we modify the test data, the trained classifier is also changed. 
  \item \textbf{Discreteness} When modifying the graph structure, the search space for adversarial example is discrete. Previous gradient methods to find adversarial examples may be invalid in this case.
\end{itemize}

 Below are the methods used by some successful works to attack and defend graph neural networks.

\subsection{Definitions for Graphs and Graph Models}

In this section, the notations and definitions of the graph structured data and graph neural network models are defined below. A graph can be represented as $G=\{\mathcal{V},\mathcal{E}\}$, where $\mathcal{V}$ is a set of $N$ nodes and $\mathcal{E}$ is a set of $M$ edges. The edges describe the connections between the nodes, which can also be expressed by an adjacency matrix ${\bf A}\in \{0,1\}^{N\times N}$. Furthermore, a graph $G$ is called an attributed graph if each node in $\mathcal{V}$ is associated with a $d$-dimensional attribute vector $x_v\in \mathbf{R}^{d}$. The attributes for all the nodes in the graph can be summarized as a matrix ${\bf X}\in {\bf R}^{N\times d}$, the $i$-th row of which represents the attribute vector for node $v_i$.

The goal of node classification is to learn a function $g: \mathcal{V}\rightarrow \mathcal{Y}$ that maps each node to one class in $\mathcal{Y}$, based on a group of labeled nodes in $G$. One of the most successful node classification models is Graph Convolutional Network (GCN) \citep{kipf2016semi}. The GCN model keeps aggregating the information from neighboring nodes to learn representations for each node $v$,
\begin{equation*}\hspace{0.2cm}
    H^{(0)} = X; ~~~
    H^{(l+1)} = \sigma(\hat{A} H^{(l)}W^l).
\end{equation*}
where $\sigma$ is a non-linear activation function, the matrix $\hat{A}$ is defined as $\hat{A}  = \tilde{D}^{-\frac{1}{2}} \tilde{A}\tilde{D}^{-\frac{1}{2}}$, $\tilde{A} = A + I_N$, and $\tilde{D}_{ii} = \sum_j \tilde{A}_{ij}$. The last layer outputs the score vectors of each node for prediction: $H^{(m)}_v = F(v,X)$. 
        
\subsection{Zugner's Greedy Method}
\begin{figure}[t]
    \centering
    \includegraphics{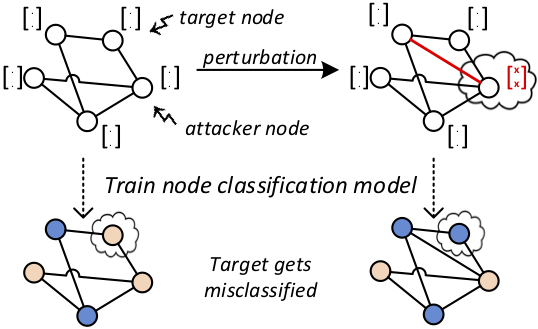}
    \caption{Adding an edge to alter the prediction of Graph Convolutional Network. (Image Credit: \citep{zugner2018adversarial})}
    \label{fig:zugner}
\end{figure}

In the work of \citep{zugner2018adversarial}, they consider attacking node classification models, Graph Convolutional Networks \citep{kipf2016semi}, by modifying the nodes connections or node features (binary). In this setting, an adversary is allowed to add/remove edges between nodes, or flip the feature of nodes with limited number of operations. The goal is to mislead the GCN model which is trained on the perturbed graph (transductive learning) to give wrong predictions. In their work, they also specify three levels of adversary capabilities: they can manipulate (1) all nodes, (2) a set of nodes $\mathcal{A}$ including the target victim $x$, and (3) a set of nodes $\mathcal{A}$ which does not include the target node $x$. A sketch is shown in Figure \ref{fig:zugner}.

Similar to the objective function in \citep{carlini2017towards} for image data, they formulate the graph attacking problem as a search for a perturbed graph $G'$ such that the learned GCN classifier $Z^*$ has the largest score margin:
\begin{equation}\label{eq:graph}
    \max\limits_{i \neq y}~~ ln(Z^*_{y}(v_0,G')) - ln (Z^*_{i}(v_0,G'))
\end{equation}
The authors solve this objective by finding perturbations on a fixed, linearized substitute GCN classifier $G_{sub}$ which is trained on the clean graph. They use a heuristic algorithm to find the most influential operations on graph $G_{sub}$ (e.g. removing/adding the edge or flipping the feature which can cause largest increase in (\ref{eq:graph})). The experimental results demonstrate that the adversarial operations are also effective on the later trained classifier $Z^*$.

During the attacking process, the authors also impose two key constraints to ensure the similarity of the perturbed graph to the original one: (1) the degree distribution should be maintained, and (2) two positive features which never happen together in $G$ should also not happen together in $G'$. Later, some other graph attacking works (e.g. \citep{ma2019attacking}) suggest the eigenvalues/eigenvectors of the graph Laplacian Matrix should also be maintained during attacking, otherwise the attacks are easily detected. However, there is still no firm consensus on how to formally define the similarity between graphs and generate unnoticeable perturbation.

\subsection{Dai's RL Method : \textit{RL-S2V}}

Different from Zugner's greedy method, the work \citep{dai2018adversarial} introduced a Reinforcement Learning method to attack the graph neural networks. This work only considers adding or removing edges to modify the graph structure. 

In \citep{dai2018adversarial}'s setting, a node classifier $F$ trained on the clean graph $G^{(0)} = G$ is given, node classifier $F$ is unknown to the attacker, and the attacker is allowed to modify $m$ edges in total to alter $F$'s prediction on the victim node $v_0$. The authors formulate this attacking mission as a Q-Learning game \citep{mnih2013playing}, with the defined Markov Decision Process as below: 
\begin{itemize}[itemsep = 0pt]
  \item \textbf{State} The state $s_t$ is represented by the tuple $(G^{(t)},v_0)$, where $G^{(t)}$ is the modified graph with $t$ iterative steps.
  \item \textbf{Action} To represent the action to add/remove edges, a single action at time step $t$ is $a_t\in \mathcal{V}\times\mathcal{V}$, which means the edge to be added or removed.
  \item \textbf{Reward} In order to encourage actions to fool the classifier, we should give a positive reward if $v_0$'s label is altered. Thus, the authors define the reward function as : $r(s_t,a_t) = 0, \forall t = 1,2,...,m-1$, and for the last step: 
  \begin{equation*}
  r(s_m, a_m) =
    \begin{cases}
      1 & \text{if} ~~C(v_0, G^{(m)}) \neq y\\
      -1 & \text{if}~~ C(v_0, G^{(m)}) = y
    \end{cases}       
  \end{equation*}
  \item \textbf{Termination} The process stops once the agent finishes modifying $m$ edges. 
\end{itemize}
The Q-Learning algorithm helps the adversary have knowledge about which actions to take (add/remove which edge) on the given state (current graph structure), in order to get largest reward (change $F$'s output).

\subsection{Graph Structure Poisoning via Meta-Learning}
Previous graph attack works only focus on attacking one single victim node. The work in~\citep{zugner2019adversarial} attempts to poison the graph so that the global node classification performance of GCN can be undermined and made almost useless. The approach is based on \textit{meta learning} \citep{finn2017model}, which is traditionally used for hyperparameter optimization, few-shot image recognition, and fast reinforcement learning. In the work \citep{zugner2019adversarial}, they use meta learning technique which takes the graph structure as the hyperparameter of the GCN model to optimize. Using their algorithm to perturb $5\%$ edges of a CITESEER graph dataset, they can increase the misclassification rate to over $30\%$.

\subsection{Attack on Node Embedding}
The work in \citep{bojcheski2018adversarial} studies how to perturb the graph structure in order to corrupt the quality of node embedding, and consequently hinder subsequent learning tasks such as node classification or link prediction. Specifically, they study DeepWalk \citep{perozzi2014deepwalk} as a random-walk based node embedding learning approach and approximately find the graph which has the largest loss of the learned node embedding.

\subsection{ReWatt: Attacking Graph Classifier via Rewiring}
The ReWatt method~\citep{ma2019attacking} attempts to attack the graph classification models, where each input of the model is a whole graph. The proposed algorithm can mislead the model by making unnoticeable perturbations on graph.

In their attacking scheme, they utilize reinforcement learning to find a rewiring operation $a=(v_1,v_2,v_3)$ at each step, which is a set of 3 nodes. The first two nodes were connected in the original graph and the edge between them is removed in the first step of the rewiring process. The second step of the rewiring process adds an edge between the node $v_1$ and $v_3$, where $v_3$ is constrained to be within $2$-hops away from $v_1$. Some analysis in~\citep{ma2019attacking} shows that the rewiring operation tends to keep the eigenvalues of the graph's Laplacian matrix, which makes it difficult to detect the attacker.

\subsection{Defending Graph Neural Networks}
Many works have shown that graph neural networks are vulnerable to adversarial examples, even though there is still no consensus on how to define the unnoticeable perturbation. Some defending works have already appeared. Many of them are inspired by the popular defense methodology in image classification, using adversarial training to protect GNN models, \citep{feng2019graph,xu2019topology}, which provides moderate robustness.
\section{Adversarial Examples in Audio and Text Data}\label{nlp}
Adversarial examples also exist in DNNs' applications in audio and text domains. An adversary can craft fake speech or fake sentences that mislead the machine language processors. Meanwhile, deep learning models on audio/text data have already been widely used in many tasks, such as Apple Siri and Amazon Echo. Therefore, the studies on adversarial examples in audio/text data domain also deserve our attention.

As for text data, the discreteness nature of the inputs makes the gradient-based attack on images not applicable anymore and forces people to craft discrete perturbations on different granularities of text (character-level, word-level, sentence-level, etc.). In this section, we introduce the related works in attacking NLP architectures for different tasks. 

\subsection{Speech Recognition Attacks}
The work in \citep{carlini2018audio} studies how to attack state-of-art speech-to-text transcription networks, such as DeepSpeech \citep{hannun2014deep}. In their setting, when given any speech waveform $x$, they can add an inaudible sound perturbation $\delta$ that makes the synthesized speech $x+\delta$ be recognized as any targeted desired phrase.
 
In their attacking work, they limited the maximum Decibels (dB) on any time of the added perturbation noise, so that the audio distortion is unnoticeable. Moreover, they inherit the C \& W's attack method \citep{carlini2017towards} on their audio attack setting.

\subsection{Text Classification Attacks}
Text classification is one of main tasks in natural language processing. In text classification, the model is devised to understand a sentence and correctly label the sentence. For example, text classification models can be applied on the IMDB dataset for characterizing user's opinion (positive or negative) on the movies, based on the provided reviews. Recent works of adversarial attacks have demonstrated that text classifiers are easily misguided by slightly modifying the texts' spelling, words or structures.

\subsubsection{Attacking Word Embedding}
The work \citep{miyato2016adversarial} considers to add perturbation on the word embedding \citep{mikolov2013distributed}, so as to fool a LSTM \citep{hochreiter1997long} classifier. However, this attack only considers perturbing the word embedding, instead of original input sentence.

\subsubsection{Attacking Words and Letters}
\begin{figure}[h]
    \centering
    \includegraphics[width=80mm]{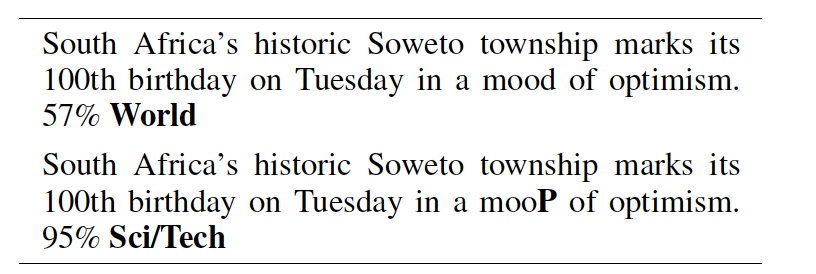}
    \caption{Replace one letter in a sentence to alter a text classifier's prediction on a sentence's topic. (Image Credit: \citep{ebrahimi2017hotflip})}
    \label{fig:hot}
\end{figure}

The work HotFlip \citep{ebrahimi2017hotflip} considers to replace a letter in a sentence in order to mislead a character-level text classifier (each letter is encoded to a vector). For example, as shown in Figure \ref{fig:hot}, changing a single letter in a sentence alters the model's prediction on its topic. The attack algorithm manages to achieve this by finding the most-influential letter replacement via gradient information. These adversarial perturbations can be noticed by human readers, but they don't change the content of the text as a whole, nor do they affect human judgments.

The work in \citep{liang2017deep} considers to manipulate the victim sentence on word and phrase levels. They try adding, removing or modifying the words and phrases in the sentences. In their approach, the first step is similar to HotFlip \citep{ebrahimi2017hotflip}. For each training sample, they find the most-influential letters, called ``hot characters''. Then, they label the words that have more than 3 ``hot characters'' as ``hot words''.  ``Hot words'' composite ``hot phrases'', which are most-influential phrases in the sentences. Manipulating these phrases is likely to influence the model's prediction, so these phrases composite a ``vocabulary'' to guide the attacking. When given a sentence, an adversary can use this vocabulary to find the weakness of the sentence, add one hot phrase, remove a hot phrase in the given sentence, or insert a meaningful fact which is composed of hot phrases.

DeepWordBug \citep{gao2018black} and TextBugger \citet{li2018textbugger} are black-box attack methods for text classification. The basic idea of the former is to define a scoring strategy to identify the key tokens which will lead to a wrong prediction of the classifier if modified. Then they try four types of ``imperceivable'' modifications on such tokens: swap, substitution, deletion and insertion, to mislead the classifier. The latter follows the same idea, and improves it by introducing new scoring functions.

The works \citep{samanta2017towards,iyyer2018adversarial} start to craft adversarial sentences that grammatically correct and maintain the syntax structure of the original sentence. The work in \citep{samanta2017towards} achieves this by using synonyms to replace original words, or adding some words which have different meanings in different contexts. On the other hand, the work \citep{iyyer2018adversarial} manages to fool the text classifier by paraphrasing the structure of sentences.

The work in \citep{DBLP:journals/corr/abs-1812-00151} conducts sentence and word paraphrasing on input texts to craft adversarial examples.
In this work, they first build a paraphrasing corpus that contains a lot of word and sentence paraphrases.
To find an optimal paraphrase of an input text, a greedy method is adopted to search valid paraphrases for each word or sentence from the corpus.
Moreover, they propose a gradient-guided method to improve the efficiency of greedy search.
This work also has significant contributions in theory: they formally define the task of discrete adversarial attack as an optimization problem on a set functions and they prove that the greedy algorithm ensures a $1-\frac{1}{e}$ approximation factor for CNN and RNN text classifiers.

\subsection{Adversarial Examples in Other NLP Tasks}
\subsubsection{Attack on Reading Comprehension Systems}
The work \citep{jia2017adversarial} studies whether Reading Comprehension models are vulnerable to adversarial attacks. In reading comprehension tasks, the machine learning model is asked to answer a given question, based on the model's ``understanding'' from a paragraph of an article. For example, the work \citep{jia2017adversarial} concentrates on Stanford Question Answering Dataset (SQuAD) where systems answer questions about paragraphs from Wikipedia. 

\begin{figure}[h!]
    \centering
    \includegraphics[width=80mm]{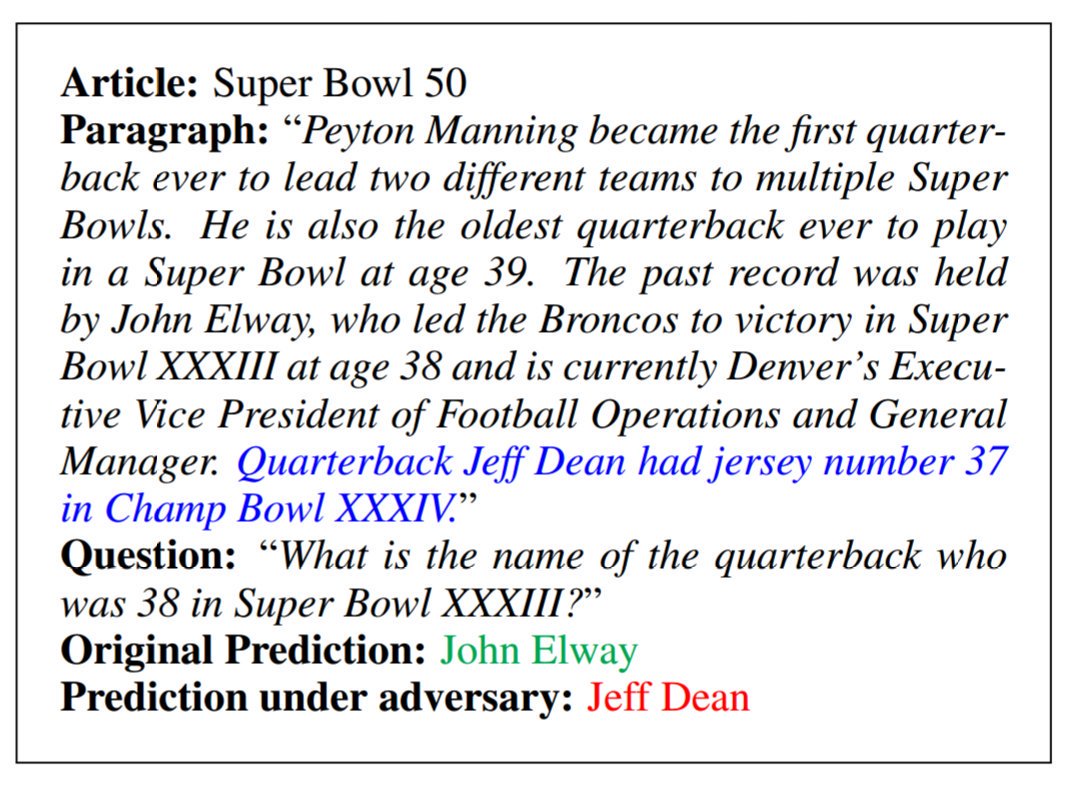}
    \caption{By adding an adversarial sentence which is similar to the answer, the reading comprehension model gives a wrong answer. (Image Credit: \citep{jia2017adversarial})}
    \label{fig:read}
\end{figure}

The authors successfully degrade the intelligence of the state-of-art reading comprehension models on SQuAD by inserting adversarial sentences. As shown in Figure \ref{fig:read}, the inserted sentence (blue) looks similar to the question, but it does not contradict the correct answer. This inserted sentence is understandable for human readers but confuses the machine a lot. As a result, the proposed attacking algorithm reduced the performance of 16 state-of-art reading comprehension models from average 75\% F1 score (accuracy) to 36\%.

Their proposed algorithm \textit{AddSent} shows a four-step operation to find adversarial sentences. 
\begin{enumerate}[itemsep = 0pt]
    \item \textit{Fake Question}: What is the name of the quarterback whose jersey number is 37 in Champ Bowl XXXIV?
    \item \textit{Fake Answer}: Jeff Dean
    \item \textit{Question to Declarative Form}: Quarterback Jeff Dean is jersey number 37 in Champ Bowl XXXIV.
    \item \textit{Grammatically Correct}: Quarterback Jeff Dean had jersey number 37 in Champ Bowl XXXIV.
\end{enumerate}

\subsubsection{Attack on Neural Machine Translation}

The work in \citep{belinkov2017synthetic} studies the stability of machine learning translation tools when their input sentences are perturbed from natural errors (typos, misspellings, etc.) and manually crafted distortions (letter replacement, letter reorder). The experimental results show that the state-of-arts translation models are vulnerable to both two types of errors, and suggest adversarial training to improve the model's robustness.

Seq2Sick \citep{cheng2018seq2sick} tries to attack seq2seq models in neural machine translation and text summarization. In their setting, two goals of attacking are set: to mislead the model to generate an output which has on overlapping with the ground truth, and to lead the model to produce an output with targeted keywords. The model is treated as a white-box and the attacking problem is formulated as an optimization problem where they seek to solve a discrete perturbation by minimizing a hinge-like loss function.

\subsection{Dialogue Generation}

Unlike the tasks above where success and failure are clearly defined, in the task of dialogue, there is no unique appropriate response for a given context. Thus, instead of misleading a well-trained model to produce incorrect outputs, works about attacking dialogue models seek to explore the property of neural dialogue models to be interfered by the perturbations on the inputs, or lead a model to output targeted responses.

The work in~\citep{niu2018adversarial} explores the over-sensitivity and over-stability of neural dialogue models by using some heuristic techniques to modify original inputs and observe the corresponding outputs.
They evaluate the robustness of dialogue models by checking whether the outputs change significantly with the modifications on the inputs. They also investigate the effects that take place when retraining the dialogue model using these adversarial examples to improve the robustness and performance of the underlying model.

In the work \citep{he2018detecting}, the authors try to find trigger inputs which can lead a neural dialogue model to generate targeted egregious responses.
They design a search-based method to determine the word in the input that maximizes the generative probability of the targeted response. Then, they treat the dialogue model as a white-box and take advantage of the gradient information to narrow the search space. Finally they show that this method works for ``normal'' targeted responses which are decoding results for some input sentences, but for manually written malicious responses, it hardly succeeds. 

The work \citep{liu2019say} treats the neural dialogue model as a black-box and adopts a reinforcement learning framework to effectively find trigger inputs for targeted responses. The black-box setting is stricter but more realistic, while the requirements for the generated responses are properly relaxed. The generated responses are expected to be semantically identical to the targeted ones but not necessarily exactly match with them.

\section{Adversarial Examples in Miscellaneous Tasks}

In this section, we summarize some adversarial attacks in other domains. Some of these domains are safety-critical, so the studies on adversarial examples in these domains are also important.

\subsection{Computer Vision Beyond Image Classification}

\begin{enumerate}
    \item \textit{Face Recognition}\\
    The work \citep{sharif2016accessorize} seeks to attack face recognition models on both a digital level and physical level. The main victim model is based on the architecture in \citep{parkhi2015deep}, which is a 39-layer DNN model for face recognition tasks. The attack on the digital level is based on traditional attacks, like Szegedy's L-BFGS method (Section \ref{sze}).
    
    Beyond digital-level adversarial faces, they also succeed in misleading face recognition models on physical level. They achieve this by asking subjects to wear their 3D printed sunglasses frames. The authors optimize the color of these glasses by attacking the model on a digital level: by considering various adversarial glasses and the most effective adversarial glasses are used for attack. As shown in Figure \ref{fig:face}, an adversary wears the adversarial glasses and successfully fool the detection of victim face recognition system.
    \begin{figure}[ht]
        \centering
        \includegraphics[width = 70mm]{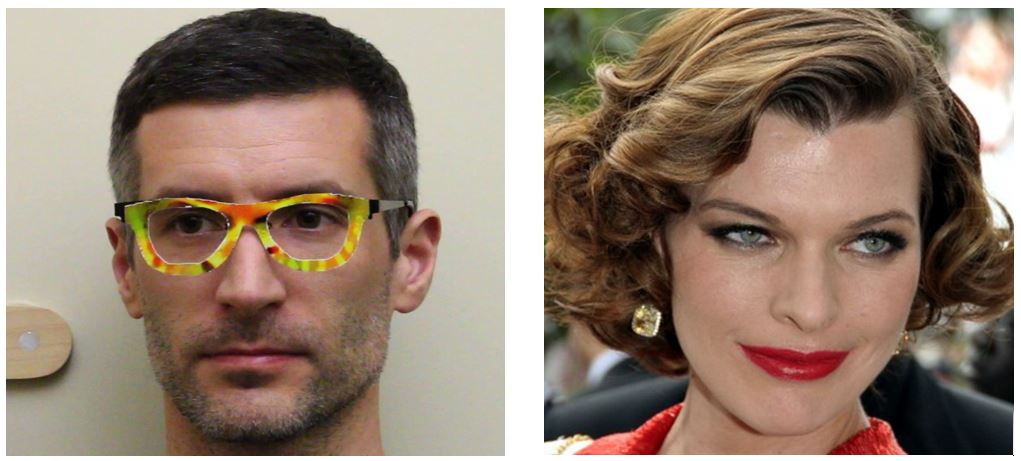}
        \caption{An adversary (left) wears a pair of adversarial glasses and is recognized as a movie-star, Milla Jovovich. (Image Credit: \citep{sharif2016accessorize})}
        \label{fig:face}
    \end{figure}
    \item \textit{Object Detection and Semantic Segmentation}\\
    There are also studies on semantic segmentation and object detection models in computer vision~\citep{xie2017adversarial,metzen2017universal}. In both semantic segmentation and object detection tasks, the goal is to learn a model that associates an input image $x$ with a series of labels $\mathcal{Y} = \{y_1,y_2,...y_N\}$. Semantic segmentation models give each pixel of $x$ a label $y_i$, so that the image is divided to different segments. Similarly, object detection models label all proposals (regions where the objects lie).
    
    The work \citep{xie2017adversarial} can generate an adversarial perturbation on $x$ which can cause the classifier to give wrong prediction on all the output labels of the model, in order to fool either semantic segmentation or object detection models. The work~\citep{metzen2017universal} finds that there exists universal perturbation for any input image for semantic segmentation models.
    
\end{enumerate}

\subsection{Video Adversarial Examples}
Most works concentrate on attacking static image classification models. However, success on image attacks cannot guarantee that there exist adversarial examples on videos and video classification systems. The work \citep{li2018adversarial} uses GAN \citep{goodfellow2014generative} to generate a dynamic perturbation on video clips that can mislead the classification of video classifiers.

\subsection{Generative Models}
The work \citep{kos2018adversarial} attacks the variational autoencoder (VAE) \citep{kingma2013auto} and VAE-GAN \citep{larsen2015autoencoding}. Both VAE and VAE-GAN use an encoder to project the input image $x$ into a lower-dimensional latent representation $z$, and a decoder to reconstruct a new image $\hat{x}$ from $z$. The reconstructed image should maintain the same semantics as the original image. In \citep{kos2018adversarial}'s attack setting, they aim to slightly perturb the input image $x$ fed to encoder, which will cause the decoder to generate image $f_{dec}(f_{enc}(x))$ having different meaning from the input $x$. For example, in MNIST dataset, the input image is "1", and the reconstructed image is "0".

\subsection{Malware Detection}
The existence of adversarial examples in safety-critical tasks, such as malware detection, should be paid much attention. The work \citep{grosse2016adversarial} built a DNN model on the DREBIN dataset by~\citep{arp2014drebin}, which contains 120,000 Android application samples, where over 5,000 are malware samples. The trained model has $97\%$ accuracy, but malware samples can evade the classifier if attackers add fake features to them. Some other works \citep{hu2017generating,anderson2016deepdga} consider using GANs \citep{goodfellow2014generative} to generate adversarial malware.

\subsection{Fingerprint Recognizer Attacks}
Fingerprint recognition systems are also one of the most safety-critical fields where machine learning models are adopted.
While, there are adversarial attacks undermining the reliability of these models. For example, fingerprint spoof attacks copy an authorized person's fingerprint and replicate it on some special materials such as liquid latex or gelatin. Traditional fingerprint recognition techniques especially minutiae-based models fail to distinguish the fingerprint images generated from different materials. The works \citep{chugh2018fingerprint,chugh2018fingerprint1}
design a modified CNN to effectively detect the fingerprint spoof attack.

\subsection{Reinforcement Learning}
\begin{figure}[t]
    \centering
    \includegraphics[width=90mm]{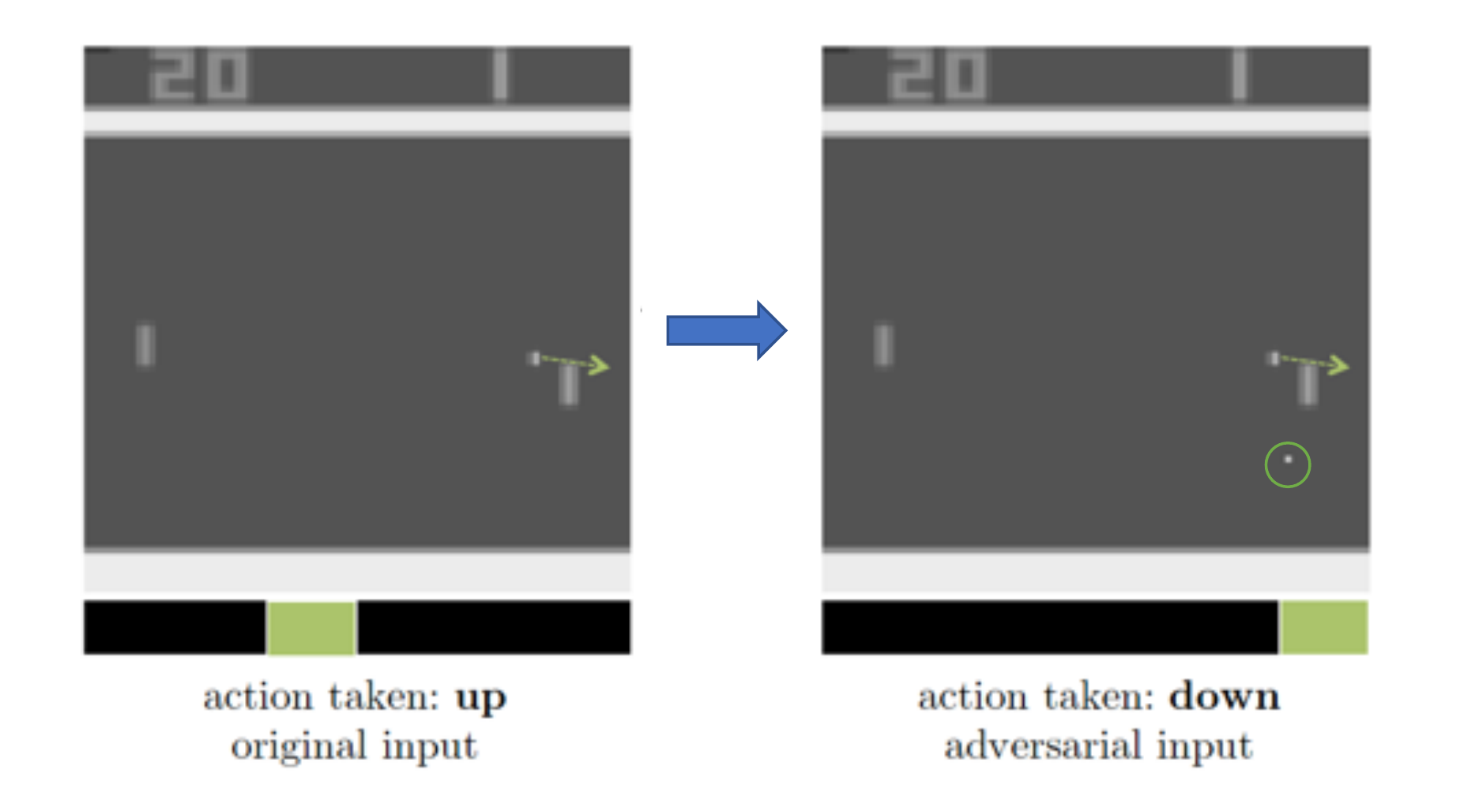}
    \caption{Left figure: the brick takes correct actions to go up to catch the ball. Right figure: the current state is perturbed by changing one pixel. The policy gives an incorrect command to go down. (Image Credit: \citep{huang2017adversarial})}
    \label{fig:pong}
\end{figure}

Different from classification tasks, deep reinforcement learning (RL) aims to learn how to perform some human tasks, such as play Atari 2600 games \citep{mnih2013playing} or play Go \citep{silver2016mastering}. For example, to play an Atari game \textit{Pong}, (Figure \ref{fig:pong}-left), the trained model takes input from the latest images of game video (state - $x$), and outputs a decision to move up or down (action - $y$). The learned model can be viewed as a rule (policy - $\pi_\theta$) to win the game (reward - $\mathcal{L}(\theta,x,y))$. A simple sketch can be: $x \xrightarrow{\pi_\theta} y$, which is in parallel to classification tasks: $x\xrightarrow{f} y$. The RL algorithms are trained to learn the parameters of $\pi_\theta$.

The work \citep{huang2017adversarial} shows that deep reinforcement learning models are also vulnerable to adversarial examples. Their approach is inherited from FGSM \citep{goodfellow2014explaining}, to take one-step gradient on the state $x$ (latest images of game video) to craft a fake state $x'$. The policy's decision on $x'$ can be totally useless to achieve the reward. Their results show that a slight perturbation on RL models' state, can cause large difference on the models' decision and performance. Their work shows that Deep Q Learning \citep{mnih2013playing}, TRPO \citep{schulman2015trust} and A3C \citep{mnih2016asynchronous} are all vulnerable to their attacks.

\section{Conclusion}
In this survey, we give a systemic, categorical and comprehensive overview on the recent works regarding adversarial examples  and their countermeasures, in multiple data domains. We summarize the studies from each section in the chronological order as shown in Appendix B, because these works are released with relatively high frequency in response to one another. The current state-of-the-art attacks will likely be neutralized by new defenses, and these defenses will subsequently be circumvented. We hope that our work can shed some light on the main ideas of adversarial learning and related applications in order to encourage progress in the field.

\section*{Acknowledgements}

This work is supported by the National Science Foundation (NSF) under grant numbers IIS-1845081 and CNS-1815636.

\small
\bibliography{example_paper}
\bibliographystyle{icml2019}


\normalsize
\onecolumn

\begin{landscape}
\appendix{Appendix}
\section{Dichotomy of Attacks}
\small

\begin{center}\label{tab:attack}
\begin{tabular}{||c|c|c|c|c|c||}
\hline
Attack & Publication & Similarity  &Attacking Capability  & Algorithm &Apply Domain  \\
\hline\hline
L-BFGS &\citep{szegedy2013intriguing} &$l_2$ &White-Box  & Iterative & Image Classification  \\
\hline
FGSM &\citep{goodfellow2014explaining} &$l_\infty$,$l_2$  &White-Box  & Single-Step & Image Classification  \\
\hline
Deepfool &\citep{moosavi2016deepfool} &$l_2$  &White-Box  & Iterative & Image Classification  \\
\hline
JSMA & \citep{papernot2016limitations} & $l_2$  &White-Box  & Iterative & Image Classification  \\
\hline
BIM & \citep{kurakin2016adversarial} & $l_\infty$  &White-Box  & Iterative & Image Classification  \\
\hline
C \& W & \citep{carlini2017towards} & $l_2$  &White-Box  & Iterative & Image Classification  \\
\hline
Ground Truth & \citep{carlini2017provably} & $l_0$  &White-Box  & SMT solver & Image Classification  \\
\hline
Spatial & \citep{xiao2018spatially} & Total Variation  &White-Box  & Iterative & Image Classification  \\
\hline
Universal & \citep{metzen2017universal} & $l_\infty$, $l_2$  &White-Box  & Iterative & Image Classification  \\
\hline
One-Pixel & \citep{su2019one} & $l_0$  &White-Box  & Iterative & Image Classification  \\
\hline
EAD & \citep{chen2018ead} & $l_1+l_2$, $l_2$  &White-Box  & Iterative & Image Classification  \\
\hline
Substitute & \citep{papernot2017practical} & $l_p$  &Black-Box  & Iterative & Image Classification  \\
\hline
ZOO & \citep{chen2017zoo} & $l_p$  &Black-Box  & Iterative & Image Classification  \\
\hline
Biggio & \citep{biggio2012poisoning} & $l_2$  &Poisoning  & Iterative & Image Classification  \\
\hline
Explanation & \citep{koh2017understanding} & $l_p$  &Poisoning & Iterative & Image Classification  \\
\hline
Zugner's & \citep{zugner2018adversarial} & Degree Distribution, Coocurrence &Poisoning  & Greedy & Node Classification  \\
\hline
Dai's & \citep{dai2018adversarial} & Edges &Black-Box  & RL & Node \& Graph Classification  \\
\hline
Meta & \citep{zugner2019adversarial} & Edges &Black-Box  & RL & Node Classification  \\
\hline
C \& W & \citep{carlini2018audio} & max dB &White-Box  & Iterative & Speech Recognition  \\
\hline
Word Embedding & \citep{miyato2016adversarial} & $l_p$ &White-Box  & One-Step & Text Classification \\
\hline
HotFlip & \citep{ebrahimi2017hotflip} & letters &White-Box  & Greedy & Text Classification \\
\hline
Jia \& Liang & \citep{jia2017adversarial} & letters &Black-Box  & Greedy & Reading Comprehension \\
\hline
Face Recognition & \citep{sharif2016accessorize} & physical &White-Box  & Iterative & Face Recognition \\
\hline
RL attack & \citep{huang2017adversarial} & $l_p$ &White-Box  & RL \\
\hline

\end{tabular}

\end{center}

\end{landscape}

\normalsize

\section{Dichotomy of Defenses}
\begin{tikzpicture}[
  level 1/.style={sibling distance=60mm},
  edge from parent/.style={->,draw},
  >=latex]

\node[root] {Defenses}
  child {node[level 2] (c1) {Gradient Masking}}
  child {node[level 2] (c2) {Robust Optimization}}
  child {node[level 2] (c3) {Detection}};

\begin{scope}[every node/.style={level 3}]
\node [below of = c1, xshift=15pt] (c11) {\textbf{Shattered Gradient}};
\node [below of = c11, xshift=15pt] (p11) {\citep{buckman2018thermometer}};
\node [below of = p11] (p12) {\citep{guo2017countering}};

\node [below of = p12, xshift=-15pt] (c12) {\textbf{Shattered Gradient}};
\node [below of = c12, xshift=15pt] (p21) {\citep{dhillon2018stochastic}};
\node [below of = p21] (p22) {\citep{xie2017mitigating}};

\node [below of = p22, xshift=-15pt] (c13) {\textbf{Exploding/Vanishing Gradient}};
\node [below of = c13, xshift=15pt] (p31) {\citep{song2017pixeldefend}};
\node [below of = p31] (p32) {\citep{samangouei2018defense}};
\node [below of = p32] (p33) {\citep{papernot2016distillation}};


\node [below of = c3, xshift=15pt] (c31) {\textbf{Auxilliary Model}};
\node [below of = c31, xshift=15pt] (k11) {\citep{grosse2016adversarial}};
\node [below of = k11] (k12) {\citep{gong2017adversarial}};
\node [below of = k12] (k13) {\citep{metzen2017detecting}};

\node [below of = k13, xshift = -15pt] (c32) {\textbf{Statistical Methods}};
\node [below of = c32, xshift=15pt] (k21) {\citep{hendrycks2016early}};
\node [below of = k21] (k22) {\citep{grosse2017statistical}};
\node [below of = k22] (k23) {\citep{gretton2012kernel}};

\node [below of = k23, xshift = -15pt] (c33) {Check Consistency};
\node [below of = c33, xshift=15pt] (k31) {\citep{feinman2017detecting}};
\node [below of = k31] (k32) {\citep{xu2017feature}};

\node [below of = c2, xshift=15pt] (c21) {\textbf{Adversarial Training}};
\node [below of = c21, xshift=15pt] (d11) {\citep{goodfellow2014explaining}};
\node [below of = d11] (d12) {\citep{madry2017towards}};
\node [below of = d12] (d13) {\citep{tramer2017ensemble}};
\node [below of = d13] (d14) {\citep{sinha2017certifying}};

\node [below of = d14, xshift = -15pt] (c22) {\textbf{Certified Defense}};
\node [below of = c22, xshift=15pt] (d21) {\citep{wong2017provable}};
\node [below of = d21] (d22) {\citep{hein2017formal}};
\node [below of = d22] (d23) {\citep{raghunathan2018certified}};
\node [below of = d23] (d24) {\citep{sinha2017certifying}};

\node [below of = d24, xshift = -15pt] (c23) {\textbf{Regularization}};
\node [below of = c23, xshift=15pt] (d31) {\citep{cisse2017parseval}};
\node [below of = d31] (d32) {\citep{gu2014towards}};
\node [below of = d32] (d33) {\citep{rifai2011contractive}};

\end{scope}

\foreach \value in {1,2,3}
  \draw[->] (c1.west) |- (c1\value.west);

\foreach \value in {1,2,3}
  \draw[->] (c2.west) |- (c2\value.west);

\foreach \value in {1,2,3}
  \draw[->] (c3.west) |- (c3\value.west);

  \draw[->] (c11.west) |- (p11.west);
  \draw[->] (c11.west) |- (p12.west);

  \draw[->] (c12.west) |- (p21.west);
  \draw[->] (c12.west) |- (p22.west);

  \draw[->] (c13.west) |- (p31.west);
  \draw[->] (c13.west) |- (p32.west);
  \draw[->] (c13.west) |- (p33.west);

  \draw[->] (c21.west) |- (d11.west);
  \draw[->] (c21.west) |- (d12.west);
  \draw[->] (c21.west) |- (d13.west);
  \draw[->] (c21.west) |- (d14.west);

  \draw[->] (c22.west) |- (d21.west);
  \draw[->] (c22.west) |- (d22.west);
  \draw[->] (c22.west) |- (d23.west);
  \draw[->] (c22.west) |- (d24.west);
  
  \draw[->] (c23.west) |- (d31.west);
  \draw[->] (c23.west) |- (d32.west);
  \draw[->] (c23.west) |- (d33.west);

  \draw[->] (c31.west) |- (k11.west);
  \draw[->] (c31.west) |- (k12.west);
  \draw[->] (c31.west) |- (k13.west);

  \draw[->] (c32.west) |- (k21.west);
  \draw[->] (c32.west) |- (k22.west);
  \draw[->] (c32.west) |- (k23.west);

  \draw[->] (c33.west) |- (k31.west);
  \draw[->] (c33.west) |- (k32.west);

\end{tikzpicture}

\end{document}